\documentclass{article}
\usepackage[final]{colm2026_conference}

\usepackage{microtype}
\usepackage{hyperref}
\usepackage{url}
\usepackage{graphicx}
\usepackage{amsmath,amssymb}
\usepackage{booktabs}
\usepackage{multirow}
\usepackage{xcolor}
\usepackage{tikz}
\usetikzlibrary{arrows.meta,positioning,fit,calc,decorations.pathreplacing}
\usepackage{pgfplots}
\pgfplotsset{compat=1.18}
\usepackage{subcaption}
\usepackage{enumitem}
\usepackage{pifont}

\usepackage{lineno}
\usepackage{placeins}
\usepackage[normalem]{ulem}

\definecolor{darkblue}{rgb}{0, 0, 0.5}
\hypersetup{colorlinks=true, citecolor=darkblue, linkcolor=darkblue, urlcolor=darkblue}

\title{TEMPER: Testing Emotional Perturbation in Quantitative \\ Reasoning}

\author{Atahan Dokme \quad Benjamin Reichman \quad Larry Heck\\
Georgia Institute of Technology\\
\texttt{\{adokme3, bzr, lheck\}@gatech.edu}
}

\begin{document}

\ifcolmsubmission
\linenumbers
\fi

\maketitle


\begin{abstract}
Large language models are trained and evaluated on quantitative reasoning tasks written in clean, emotionally neutral language. However, real-world queries are often wrapped in frustration, urgency or enthusiasm. Does emotional framing alone degrade reasoning when all numerical content is preserved? To investigate this, a controlled emotion translation framework is developed that rewrites problems into emotional variants while preserving all quantities and relationships. Using this framework, \textsc{Temper-5400} (5,400 semantically verified emotion--neutral pairs) is constructed across GSM8K, MultiArith, and ARC-Challenge, and evaluated on eighteen models (1B to frontier scale). Two core results emerge: First, emotional framing reduces accuracy by 2--10 percentage points even though all numerical content is preserved. Second, neutralizing emotional variants recovers most of the lost performance, showing both that the degradation is tied to emotional style rather than content corruption and that neutralization can serve as a lightweight inference-time mitigation. Non-emotional paraphrases cause no such degradation, implicating emotional content rather than surface-level changes. Beyond emotion specifically, the benchmark construction procedure offers a controllable instrument for stylistic translation and robustness evaluation.
\end{abstract}


\section{Introduction}

When a frustrated student types \emph{``Ugh, my teacher is such a jerk! I got a 70 on the first exam and a lousy 45 on the second, my homework average is 90, and each component is worth 25\% of the grade. He says I need at least a 70 overall to pass this garbage class. What the hell do I need on the final to not fail?!''}, the underlying arithmetic is straightforward, yet the query looks nothing like the sanitized word problems in GSM8K \citep{cobbe2021gsm8k} or MultiArith \citep{roy2015multiarith}. Real users frame questions with human emotions but benchmarks assume emotionally neutral input. Does this mismatch matter? 

There is reason to suspect this mismatch does matter. LLM performance is fragile under surface-level perturbations such as formatting \citep{sclar2024quantifying}, irrelevant context \citep{shi2023irrelevant}, and few-shot ordering \citep{lu2022fantastically}. On the other hand, emotional signals are encoded across model layers \citep{reichman2026emotions}, suggesting they could interfere with downstream tasks. Prior studies change both tone and content simultaneously and do not evaluate the effects of human-like emotional tone on quantitative reasoning performance. This paper isolates emotion as a variable and provides numerical, behavioral and representational evidence that standard neutral benchmarks overestimate reasoning robustness under emotionally framed inputs. A teacher-student framework is developed for training controlled emotion translators that rewrite math problems into emotionally framed variants designed to preserve all numbers, quantifiers, and relationships (98--100\% measured preservation, \S\ref{sec:fidelity}). To test whether the resulting degradation is tied to emotional style rather than content corruption, the same translators can neutralize emotional variants back to neutral tone. Across three benchmarks and eighteen reasoning models (1B to frontier scale), two core findings are established: (1)~emotional framing of mathematically identical problems significantly degrades reasoning accuracy (2--10 percentage points), and (2)~neutralization acts as a partial denoising step, recovering most of the lost performance and confirming that the degradation is tied to emotional style rather than content corruption. Hidden-state analysis across four architectures shows that emotional text shifts final-layer representations 3--4$\times$ further from the original in cosine distance compared to neutralized and paraphrased versions, consistent with representational interference rather than simple rewording.

The contributions of this work are: (1)~a controllable bidirectional emotion translation instrument that combines emotional latent space distillation with intensity controllable auxiliary losses, trained on DeepSeek-V3 generated rewrite pairs; (2)~\textsc{Temper-5400}, a benchmark of 5{,}400 verified emotion--neutral pairs (900 problems $\times$ 6 emotions) across three reasoning domains, constructed using multi-translator generation, semantic filtering, and non-emotional paraphrase controls; (3)~evidence across eighteen models that emotional framing degrades reasoning by 2--10 percentage points while non-emotional paraphrases do not; (4)~evidence that neutralization recovers most lost performance, serving as both a diagnostic tool and a lightweight inference-time mitigation; and (5)~representational evidence that emotion shifts hidden states 3--4$\times$ further than neutral text with linear separability.


\section{Related work}

\paragraph{Emotion and LLM Behavior.}EmotionPrompt \citep{li2024emotionprompt} and NegativePrompt \citep{wang2024negativeprompt} show that emotional phrases alter LLM performance, and \citet{vinay2024emotional} find they amplify disinformation. Related work on toxicity \citep{deshpande2023toxicity} and politeness \citep{rao2018dear} explores how social register affects generation quality but does not measure reasoning impact. This isolation is provided through controlled translators that preserve all numerical content and evaluation across eighteen models from 1B to frontier scale.

\paragraph{Robustness of mathematical reasoning.}
Behavioral testing frameworks such as CheckList \citep{ribeiro2020checklist} systematically probe model capabilities through controlled perturbations. LLM math performance is sensitive to paraphrasing, irrelevant context \citep{shi2023irrelevant}, and prompt formatting \citep{sclar2024quantifying}, while chain-of-thought prompting \citep{wei2022chain} partially mitigates such fragilities. Concurrent work perturbs the mathematical content itself: GSM-Symbolic \citep{mirzadeh2024gsmsymbolic} changes numerical values, GSM-PLUS \citep{li2024gsmplus} augments problem structure. This work complements these by holding math constant and perturbing only the emotional wrapper, producing consistent 2--10\% drops that persist even under CoT prompting.

\paragraph{Knowledge distillation and style transfer.}
Text style transfer aims to change stylistic attributes while preserving content \citep{jin2022deep}, with sentiment and formality transfer being the most studied directions. The proposed translator extends this paradigm to emotion in mathematical text, where content preservation is verifiable through numerical constraints. Architecturally, knowledge distillation \citep{hinton2015distilling} is adapted with intermediate representation matching inspired by FitNets \citep{romero2015fitnets}: the student's hidden states are projected into the teacher's latent space providing richer supervision.


\section{Methodology}

This section describes how controlled emotion translators are built that can rewrite math problems into emotional variants while preserving mathematical content. The core idea is to combine a standard language model fine-tuned for translation with a frozen emotion classifier that provides auxiliary supervision, aligning the translator's internal representations with the classifier's emotion space. This produces translators whose emotional intensity can be controlled via a single hyperparameter. Figure~\ref{fig:pipeline} illustrates the full pipeline, with implementation details in Appendix~\ref{app:implementation}.

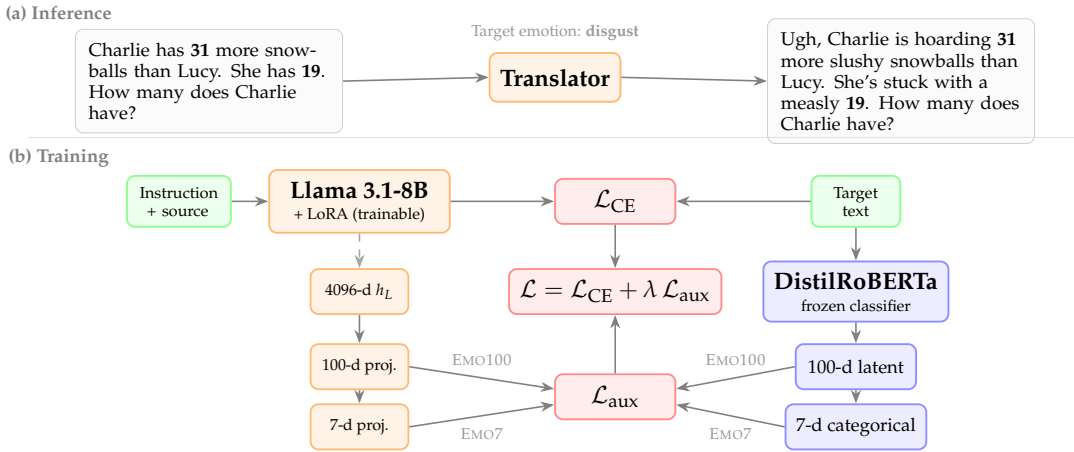
\begin{figure*}[t]
\centering
\begin{tikzpicture}[
    every node/.style={font=\footnotesize},
    box/.style={rectangle, rounded corners=3pt, minimum height=0.6cm, align=center,
                inner sep=4pt, line width=0.7pt},
    train/.style={box, fill=orange!10, draw=orange!55},
    froze/.style={box, fill=blue!7, draw=blue!45},
    loss/.style={box, fill=red!7, draw=red!45, minimum width=1.6cm},
    data/.style={box, fill=green!7, draw=green!45},
    exbox/.style={rectangle, rounded corners=3pt, draw=black!18, fill=gray!3,
                  inner sep=5pt, text width=3.2cm, align=left, font=\scriptsize},
    arr/.style={-{Stealth[length=1.8mm]}, line width=0.6pt, color=black!50},
    darr/.style={-{Stealth[length=1.8mm]}, line width=0.6pt, dashed, color=black!35},
    lbl/.style={font=\tiny, text=black!40},
]

\node[font=\scriptsize\bfseries, text=black!50] at (-4.4, 1.7) {(a) Inference};

\node[exbox] (neut) at (-2.4, 0.8) {%
Charlie has \textbf{31} more snowballs than Lucy.
She has \textbf{19}. How many does Charlie have?};

\node[train, minimum width=1.6cm, minimum height=0.65cm] (trans) at (2.2, 0.85) {\textbf{Translator}};
\node[lbl, align=center] at (2.2, 1.45) {Target emotion: \textbf{disgust}};
\draw[arr] (neut.east) -- (trans.west);

\node[exbox] (emot) at (6.8, 0.8) {%
Ugh, Charlie is hoarding \textbf{31} more
slushy snowballs than Lucy.
She's stuck with a measly \textbf{19}.
How many does Charlie have?};
\draw[arr] (trans.east) -- (emot.west);

\draw[black!12, line width=0.4pt] (-4.8, -0.25) -- (9.2, -0.25);

\node[font=\scriptsize\bfseries, text=black!50] at (-4.4, -0.48) {(b) Training};

\node[data, font=\tiny, text width=1.1cm, align=center] (inp) at (-2.8, -0.8) {Instruction\\+ source};
\node[train, minimum width=2.4cm, minimum height=0.8cm] (stu) at (-0.4, -0.8) {%
\textbf{Llama 3.1-8B}\\[-2pt]{\tiny + LoRA (trainable)}};
\draw[arr] (inp) -- (stu);

\node[data, font=\tiny, text width=0.9cm, align=center] (tgt) at (6.2, -0.8) {Target\\text};
\node[froze, minimum width=2.4cm, minimum height=0.8cm] (teacher) at (6.2, -2.0) {%
\textbf{DistilRoBERTa}\\[-2pt]{\tiny frozen classifier}};
\draw[arr] (tgt) -- (teacher);

\node[froze, minimum width=1.6cm, font=\scriptsize] (lat) at (6.2, -3.0) {100-d latent};
\node[froze, minimum width=1.6cm, font=\scriptsize] (cat) at (6.2, -3.8) {7-d categorical};
\draw[arr] (teacher.south) -- (lat.north);
\draw[arr] (lat.south) -- (cat.north);

\node[loss] (lce) at (3.0, -0.8) {$\mathcal{L}_{\text{CE}}$};
\draw[arr] (stu.east) -- (lce.west);
\draw[arr] (tgt.west) -- (lce.east);

\node[loss, minimum width=2.8cm] (total) at (3.0, -2.0) {%
$\mathcal{L} = \mathcal{L}_{\text{CE}} + \lambda\,\mathcal{L}_{\text{aux}}$};
\draw[arr] (lce) -- (total);

\node[loss] (laux) at (3.0, -3.4) {$\mathcal{L}_{\text{aux}}$};
\draw[arr] (laux) -- (total);

\node[train, minimum width=1.3cm, font=\tiny] (hid) at (-0.4, -2.0) {4096-d $h_L$};
\draw[darr] (stu.south) -- (hid.north);

\node[train, minimum width=1.3cm, font=\tiny] (sproj100) at (-0.4, -3.0) {100-d proj.};
\node[train, minimum width=1.3cm, font=\tiny] (sproj7) at (-0.4, -3.8) {7-d proj.};
\draw[arr] (hid.south) -- (sproj100.north);
\draw[arr] (sproj100.south) -- (sproj7.north);

\draw[arr] (sproj100.east) -- ([yshift=3pt]laux.west)
    node[lbl, midway, above=1pt] {\textsc{Emo100}};
\draw[arr] (sproj7.east) -- ([yshift=-3pt]laux.west)
    node[lbl, midway, below=1pt] {\textsc{Emo7}};

\draw[arr] (lat.west) -- ([yshift=3pt]laux.east)
    node[lbl, midway, above=1pt] {\textsc{Emo100}};
\draw[arr] (cat.west) -- ([yshift=-3pt]laux.east)
    node[lbl, midway, below=1pt] {\textsc{Emo7}};

\end{tikzpicture}
\caption{\textbf{Overview.} \emph{(a)}~The translator rewrites a neutral math problem into an emotional variant, preserving numerical structure (bold). \emph{(b)}~Training combines generation loss ($\mathcal{L}_{\text{CE}}$) with auxiliary alignment ($\mathcal{L}_{\text{aux}}$). $h_L$ denotes the final-layer (layer $L{=}32$) hidden state of Llama~3.1-8B (4096-d), mean-pooled and linearly projected to match the teacher's representation space. \textsc{Emo100} supervises at the 100-dim latent layer; \textsc{Emo7} at the 7-dim categorical output. $\lambda$ controls the emotion--fidelity trade-off.}
\label{fig:pipeline}
\end{figure*}

\subsection{Emotion teacher and student translator}

The emotion translator is trained using a student-teacher framework. The student (Llama~3.1-8B-Instruct with LoRA) learns to translate between neutral and emotional text,  while a frozen emotion classifier (the teacher) provides auxiliary supervision that guides the student's representations toward the desired emotional target. This setup lets the student learn from the generation loss while the teacher ensures the output carries the correct emotion. The teacher is a DistilRoBERTa classifier pretrained on GoEmotions \citep{demszky2020goemotions} for 7-class emotion classification (anger, disgust, fear, joy, neutral, sadness, surprise). On top of its frozen [CLS] hidden state, a two-layer bottleneck head is trained (768→100→7) on 72K emotional math samples (Appendix~\ref{app:implementation}), making the teacher more aligned with the target domain. This head provides two supervision signals at different granularities: (1)~the 7-class probability distribution at the output layer, which captures which emotion is expressed, and (2)~the 100-dimensional intermediate representation, which additionally encodes intensity and style. The student is Llama~3.1-8B-Instruct adapted with LoRA \citep{hu2022lora}. Final-layer hidden states are mean-pooled and linearly projected via a trainable linear layer into the teacher's representation space ($D{=}7$ or $D{=}100$, depending on variant).

\subsection{Training objective and translator variants}
\label{sec:variants}

The student is trained with a combined loss: a  cross-entropy loss 
$\mathcal{L}_{\text{CE}}$ for fluent generation, plus an auxiliary loss that 
encourages the student's representations to match the teacher's emotion space:
\begin{equation}
    \mathcal{L} = \mathcal{L}_{\text{CE}} + \lambda \cdot \mathcal{L}_{\text{aux}}
\end{equation}
Two forms of the auxiliary loss are explored, corresponding to the two teacher signals. When matching the teacher's 7-class output (\textsc{Emo7}), $\mathcal{L}_{\text{aux}}$ is a temperature-scaled KL divergence between the student's projected distribution and the teacher's class probabilities. When 
matching the teacher's 100-dimensional intermediate representation 
(\textsc{Emo100}), $\mathcal{L}_{\text{aux}}$ is the MSE between the student's projection and the teacher's latent vector. The 100-dim representation captures finer-grained information (intensity and stylistic variation within each emotion category) compared to the 7-class distribution, which only encodes categories. 

The auxiliary loss weight $\lambda$ is a training-time hyperparameter controlling the trade-off between auxiliary emotion alignment and generation fidelity. As $\lambda$ increases, the model prioritizes matching the teacher's emotional representation at the cost of fluent and semantically faithful generation. Table~\ref{tab:lambda_gradient} quantifies this gradient. For emotion, the teacher classifier provides emotion classification accuracy (Emo\%) and the pretrained classifier provides a confidence score as an independent intensity signal. Since emotion accuracy remains high ($>$93\%) across all $\lambda$, classifier confidence is used as a more discriminative intensity metric. On top of this, Amazon Mechanical Turk (AMT) annotators and LLM-as-a-judge (Claude Haiku~4.5) independently rate intensity on a 1--10 scale. For math fidelity, Math Value Preservation (MVP) is an automated check that verifies all numbers and quantitative terms from the original appear in the translation. The annotators and LLM-as-a-judge evaluate mathematical equivalence to validate MVP. Automated metrics (Emo\%, Conf, MVP) are computed on 600 translations per $\lambda$ (100 problems $\times$ 6 emotions); human and LLM evaluations use a 240-sample subset (20 per $\lambda$). The three evaluation methods form a complementary hierarchy: automated metrics provide fast surface-level checks at scale, the LLM-as-a-judge and human annotators validate them on a smaller sample size. 

\begin{table}[t]
\centering
\scriptsize
\setlength{\tabcolsep}{2pt}
\renewcommand{\arraystretch}{0.92}
\begin{tabular}{ll@{\hskip 4pt}rrrr@{\hskip 6pt}rrr}
\toprule
& & \multicolumn{4}{c}{\textbf{Emotion}} & \multicolumn{3}{c}{\textbf{Math Fidelity}} \\
\cmidrule(lr){3-6} \cmidrule(lr){7-9}
\textbf{Type} & $\boldsymbol{\lambda}$ & \textbf{Emo\%} & \textbf{Conf} & \textbf{H.Int} & \textbf{LLM.Int} & \textbf{MVP\%} & \textbf{Haiku\%} & \textbf{Human\%} \\
\midrule
\textsc{Emo100} & 0.02 & \textbf{100.0} & .795 & 5.6 & 5.8 & \textbf{98.7} & 100 & 100 \\
 & 0.2 & 99.5 & .815 & 5.7 & 5.5 & 97.3 & 100 & 100 \\
 & 1 & 97.7 & .810 & 5.6 & 5.2 & 97.7 & 90 & 98 \\
 & 4 & 95.8 & .799 & 5.0 & 5.2 & 98.0 & 100 & 98 \\
 & 100 & 98.5 & .945 & 6.7 & 6.0 & 66.5 & 10 & 83 \\
 & 400 & 99.8 & \textbf{.970} & \textbf{8.1} & \textbf{8.3} & \textbf{0.2} & 0 & 38 \\
\midrule
\textsc{Emo7} & 0.05 & 98.3 & .820 & 6.2 & 5.9 & 98.2 & 100 & 100 \\
 & 0.5 & 99.3 & .813 & 6.1 & 5.9 & 98.0 & 95 & 98 \\
 & 2.5 & 98.7 & .802 & 5.6 & 5.8 & 98.0 & 100 & 100 \\
 & 10 & 99.0 & .799 & 6.2 & 6.8 & 97.5 & 100 & 97 \\
 & 250 & 94.3 & .822 & 6.3 & 6.5 & 95.5 & 70 & 93 \\
 & 1000 & 93.3 & .877 & 6.9 & 6.9 & 83.7 & 50 & 88 \\
\bottomrule
\end{tabular}
\caption{\textbf{Intensity--fidelity trade-off} (100 GSM8K $\times$ 6 emotions, single-attempt without retries). \emph{Emotion:} Emo\%: teacher emotion accuracy. Conf: pretrained classifier confidence. H.Int / LLM.Int: human and LLM intensity (1--10). \emph{Math:} MVP\%: automated (600 samples). Haiku\% / Human\%: LLM judge and annotators (same 20 samples per $\lambda$).}
\label{tab:lambda_gradient}
\end{table}

\begin{figure}[t]
\centering
\includegraphics[width=0.72\textwidth]{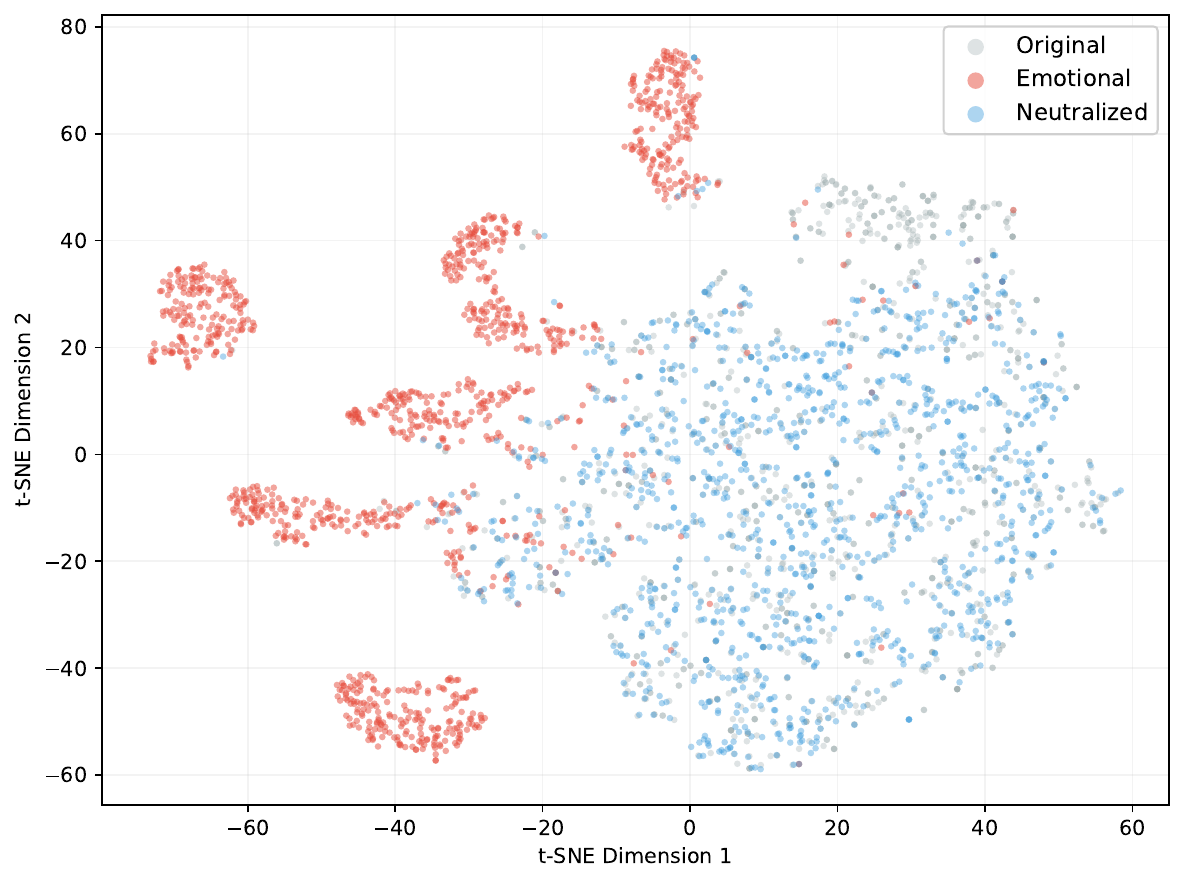}
\caption{\textbf{Original vs.\ emotional vs.\ neutralized} in the teacher's 100-dim latent space (\textsc{Emo100-L}, 1,200 GSM8K samples). Emotional translations (red) form separated clusters by emotion, while original (gray) and neutralized (blue) variants overlap in one diffuse central region.}
\label{fig:tsne_orig_emo_neu}
\end{figure}

Under \textsc{Emo100}, increasing $\lambda$ produces a smooth intensity--fidelity gradient: classifier confidence rises from .80 ($\lambda{=}0.02$) to .97 ($\lambda{=}400$). Math preservation is 98--100\% at conservative $\lambda$ across all three measures, degrading sharply at extreme weights (MVP 0.2\%, Haiku 0\%, Human 38\% at $\lambda{=}400$). Human and LLM-as-a-judge intensity ratings are flat at 5--6 for conservative $\lambda$, rising at extremes, confirming that classifier confidence is a valid metric. Qualitative examples appear in Appendix~\ref{app:intensity_example}. \textsc{Emo7} shows a much weaker trend across the same range, as seven discrete categories lack capacity for fine-grained intensity control compared to a richer 100-dim latent space. Figure~\ref{fig:tsne_orig_emo_neu} visualizes this latent space: emotional translations form separated clusters per emotion while original and neutralized variants overlap in one central region, which is the structure the auxiliary loss aligns the translator to.

 Conservative operating points are deliberately selected ($\lambda{=}0.02$ for \textsc{Emo100-L} and $\lambda{=}0.05$ for \textsc{Emo7-L}) that remain in the high-fidelity regime. The auxiliary loss weight acts as an intensity knob rather than a direct driver of degradation. At the conservative weights used in the main results all five variants produce similar drops. Intermediate \textsc{Emo100} variants trained at higher weights show the math faithful drop scaling from 2.8 to 17.5 points while the rate of genuinely broken rewrites rises sharply, which justifies the conservative operating point (Appendix~\ref{app:lambda_curve}). Separate variants are trained under each supervision type which produces greater translator diversity for the ensemble construction in \S\ref{sec:temper_construction}. Five variants are created by crossing the two supervision types with two weight levels and the CE-Only: \textbf{CE-Only} ($\lambda{=}0$), \textbf{Emo7-L} ($\lambda{=}0.05$), \textbf{Emo7-H} ($\lambda{=}0.5$), \textbf{Emo100-L} ($\lambda{=}0.02$), and \textbf{Emo100-H} ($\lambda{=}0.2$). All five variants share a single base generator (Llama~3.1-8B) and a single training data source, differing only in auxiliary loss type and weight. This design choice is intentional: holding the generator architecture fixed isolates emotion as the perturbation variable. Despite the shared generator, the variants produce semantically different translations for the same input, providing useful ensemble diversity while preserving experimental control. Additionally, a quality-control pipeline is used (Appendix~\ref{app:implementation}) during inference with up to three retries. 
\section{Experimental setup and evaluation}
\label{sec:setup}
A math corpus of 10,000 problems is sampled from AQuA-RAT \citep{ling2017aqua}, MathQA \citep{amini2019mathqa}, and ASDiv \citep{miao2020asdiv}. DeepSeek-V3 \citep{deepseek2024v3} generates emotional variants in the six Ekman basic emotions \citep{ekman1992basic}, producing 126K training pairs (60K forward, 60K backward, 6K identity). Translation quality is evaluated on GSM8K \citep{cobbe2021gsm8k}, which is disjoint from the training sources (7,914 translations per model). Full implementation details and hyperparameters are provided in Appendix~\ref{app:implementation}.

Using the three evaluation methods (Automated metrics, Human Evaluation and LLM-as-a-judge) introduced in \S\ref{sec:variants}, the five translator variants are compared across automated metrics, human evaluation, and LLM-as-a-judge validation (Table~\ref{tab:quality_merged}). Human evaluation is performed on AMT based on three criteria: emotion detection, emotional richness, and mathematical preservation (322 HITs, three annotators each; procedure in Appendix~\ref{app:mturk}). Results are reported by majority vote (Table~\ref{tab:quality_merged}b). Emotion detection reaches 86.9\% accuracy, with anger being the lowest (55\%) due to asymmetric confusion with disgust (Table~\ref{tab:confusion_matrix} in Appendix~\ref{app:confusion}). Emotional richness ratings cluster tightly (5.87--6.12/10), and mathematical preservation is near-perfect (100\% majority vote; 98.3\% individual). Annotator and LLM evaluations show strong agreement: emotion detection accuracy is comparable (86.9\% annotators vs.\ 89.3\% LLM on the same 140-problem set), and both rank the five translators identically by richness (Table~\ref{tab:quality_merged}c). The fine-tuned emotion teacher classifier achieves 99.3\% on the same samples. In the 15 cases where annotators and classifier disagree (mostly anger--disgust confusion), the classifier matches the target emotion in 14 (Appendix~\ref{app:confusion}), validating its use in the training pipeline as a teacher. All five Llama-based translator variants achieve nearly identical automated scores across emotion accuracy, BERTScore, Roundtrip BERTScore (for neutralization quality) \citep{zhang2020bertscore} and MVP (Table~\ref{tab:quality_merged}a), indicating the low-$\lambda$ auxiliary losses introduce stylistic diversity rather than altering quality.\label{sec:fidelity} 

Automated MVP checks catch surface-level numerical value changes. Claude Haiku~4.5 is used as a conservative second-pass judge, followed by manual human review of all flagged cases. The judge is deliberately conservative: of the two representative cases in Appendix~\ref{app:confusion}, one caught a genuine semantic break that annotators missed (filtered from the benchmark), while the other flagged an ambiguity already present in the original problem. After this three-stage filtering at full scale, emotionalization alters mathematical content in 0.8--1.4\% of cases; neutralization in 2.0--3.0\%. Flagged pairs are substituted with alternatives from other translator variants, as detailed in \S\ref{sec:reasoning}.

\begin{table*}[t]
\centering
\scriptsize
\setlength{\tabcolsep}{3pt}
\renewcommand{\arraystretch}{0.92}
\begin{tabular}{@{}lcccc@{\hskip 8pt}ccccc@{\hskip 8pt}cccc@{}}
\toprule
& \multicolumn{4}{c}{\textbf{(a) Automated Metrics}} & \multicolumn{5}{c}{\textbf{(b) Human Evaluation (MTurk)}} & \multicolumn{4}{c}{\textbf{(c) LLM Judge (Haiku 4.5)}} \\
\cmidrule(lr){2-5} \cmidrule(lr){6-10} \cmidrule(lr){11-14}
\textbf{Model} & \textbf{Emo} & \textbf{BS} & \textbf{RT BS} & \textbf{MVP} &
\textbf{Emo\%} & \textbf{Rich.} & \textbf{Rank} & \textbf{Top-1} & \textbf{Math\%} &
\textbf{Emo\%} & \textbf{Rank} & \textbf{Top-1} & \textbf{Math\%} \\
\midrule
\multicolumn{14}{l}{\textit{Trained translators (Llama-3.1-8B)}} \\
\textsc{CE-Only}   & .991 & .913 & .948 & .990 & 83.3 & 5.87 & 3.09 & 17.5\% & 100 & 89.3 & 2.98 & 17.1\% & \textbf{100} \\
\textsc{Emo7-L}    & .993 & .909 & .944 & .981 & \textbf{95.7} & 6.08 & 2.95 & \textbf{26.2}\% & 100 & \textbf{96.4} & 2.95 & 14.6\% & {\textbf{100}} \\
\textsc{Emo7-H}    & .994 & .906 & .941 & .983 & 92.3 & 6.07 & \textbf{2.86} & 23.0\% & 100 & 85.7 & \textbf{2.80} & 22.0\% & 96.4 \\
\textsc{Emo100-L}  & .996 & .907 & .942 & .989 & 84.0 & 6.06 & 3.20 & 11.1\% & 100 & 92.9 & 3.34 & 14.6\% & {\textbf{100}} \\
\textsc{Emo100-H}  & .980 & .909 & .945 & .986 & 79.2 & \textbf{6.12} & 2.90 & 22.2\% & 100 & 82.1 & 2.93 & \textbf{31.7\%} & {92.9} \\
\bottomrule
\end{tabular}
\caption{\textbf{Translation quality.} \textbf{(a)}~Automated: emotion accuracy, BERTScore \citep{zhang2020bertscore}, Roundtrip BERTScore, math value preservation (150 problems $\times$ 6 emotions, 3 retries). \textbf{(b)}~MTurk majority vote (322 HITs, 3 annotators each). \textbf{(c)}~Claude Haiku~4.5 ($n{=}28$ per variant); Haiku serves as a conservative filter applied at full scale in \S\ref{sec:fidelity}. Multi-architecture and baseline results in Table~\ref{tab:multi_arch_quality}.}
\label{tab:quality_merged}
\label{tab:human_eval}
\label{tab:translation_quality}
\end{table*}

{\paragraph{Multi-architecture validation and baselines.} To confirm the pipeline generalizes beyond Llama, CE-Only and Emo100-H translators are trained on Qwen-2.5-7B-Instruct and Mistral-7B-Instruct-v0.3 with identical data and hyperparameters. They achieve comparable quality (Table~\ref{tab:multi_arch_quality} in Appendix~\ref{app:baseline_examples}), demonstrating that the approach is not architecture-specific. Two untrained baselines (Llama-8B, Llama-70B) and a Llama-8B fine-tuned on non-quantitative emotional QA passages \citep{reichman2025emotional} were also tested. They all leak answers, fail to modify tone, or corrupt numbers (examples in Appendix~\ref{app:baseline_examples}).}

%

\section{Does emotion break reasoning?}
\label{sec:reasoning}
\label{sec:temper}
\label{sec:temper_construction}
\label{sec:ensemble}

All five translator variants mentioned in \S\ref{sec:variants} differ in auxiliary loss, producing different phrasings. For each problem, semantically altered candidates from these variants are filtered. Verified candidates cluster within 1--3\% BERTScore of each other, and the translation with the lowest BERTScore (greatest emotional transformation preserving mathematical content) is selected, paired with its neutralization from the same variant. Thus, the ensemble of five translator variants ensures that every emotional translation and its neutralization in the final benchmark preserves the mathematical content of the original problem.

The ensemble pipeline is applied to GSM8K \citep{cobbe2021gsm8k} (400 problems), MultiArith \citep{roy2015multiarith} (300), and ARC-Challenge \citep{clark2018arc} (200), yielding 900 problems $\times$ 6 emotions = 5{,}400 emotion--neutral pairs (10{,}800 translations total). GSM8K and MultiArith test multi-step arithmetic at different difficulty levels; ARC-Challenge provides scientific reasoning in multiple-choice format. Translator selection is approximately uniform across variants (${\sim}$19--24\% each) demonstrating the strength of the ensemble method. \textsc{Temper-5400} is publicly available.\footnote{Dataset: \url{https://huggingface.co/datasets/atahandokme/temper-5400}. Code: \url{https://github.com/atahandokme/temper-colm2026}.} Eighteen instruction-tuned reasoning models are evaluated (Table~\ref{tab:temper_results}) under base prompting and zero-shot chain-of-thought (CoT) \citep{wei2022chain}, with greedy decoding (temperature~0). Prompt templates appear in Appendix~\ref{app:prompts}. Few-shot CoT provides no additional benefit for instruction-tuned models (Appendix~\ref{app:fewshot_cot}).
\label{sec:multimodel}

\begin{table*}[t]
\centering
\tiny
\setlength{\tabcolsep}{2.5pt}
\renewcommand{\arraystretch}{0.85}
\begin{tabular}{ll cccccc cccccc cccccc}
\toprule
& & \multicolumn{6}{c}{\textbf{GSM8K}} & \multicolumn{6}{c}{\textbf{MultiArith}} & \multicolumn{6}{c}{\textbf{ARC}} \\
\cmidrule(lr){3-8} \cmidrule(lr){9-14} \cmidrule(lr){15-20}
& & \multicolumn{3}{c}{\textit{Base}} & \multicolumn{3}{c}{\textit{CoT}} & \multicolumn{3}{c}{\textit{Base}} & \multicolumn{3}{c}{\textit{CoT}} & \multicolumn{3}{c}{\textit{Base}} & \multicolumn{3}{c}{\textit{CoT}} \\
\cmidrule(lr){3-5} \cmidrule(lr){6-8} \cmidrule(lr){9-11} \cmidrule(lr){12-14} \cmidrule(lr){15-17} \cmidrule(lr){18-20}
\textbf{Tier} & \textbf{Model} & \textbf{O} & \textbf{E} & \textbf{N} & \textbf{O} & \textbf{E} & \textbf{N} & \textbf{O} & \textbf{E} & \textbf{N} & \textbf{O} & \textbf{E} & \textbf{N} & \textbf{O} & \textbf{E} & \textbf{N} & \textbf{O} & \textbf{E} & \textbf{N} \\
\midrule
\multirow{4}{*}{\rotatebox[origin=c]{90}{\textit{Tiny}}}
& Llama-3.2-1B       & 35.2 & 29.0 & 36.8 & 34.0 & 33.4 & 39.7 & 49.0 & 41.4 & 51.3 & 60.0 & 56.6 & 63.8 & 40.5 & 36.8 & 40.2 & 51.5 & 49.4 & 53.6 \\
& Gemma-2-2B         & 60.8 & 56.9 & 61.7 & 64.0 & 59.2 & 62.5 & 92.0 & 88.4 & 92.4 & 89.7 & 91.1 & 93.3 & 77.5 & 69.6 & 75.3 & 81.5 & 75.3 & 81.3 \\
& Llama-3.2-3B       & 74.5 & 64.5 & 71.5 & 76.2 & 67.6 & 74.2 & 94.0 & 83.0 & 92.0 & 97.0 & 88.4 & 96.9 & 76.5 & 67.7 & 73.4 & 80.5 & 76.4 & 79.5 \\
& Qwen-2.5-3B        & 86.2 & 79.1 & 83.9 & 86.0 & 81.1 & 85.0 & 95.0 & 92.4 & 94.1 & 97.7 & 95.0 & 97.3 & 83.0 & 75.8 & 83.8 & 86.5 & 80.5 & 88.0 \\
\midrule
\multirow{4}{*}{\rotatebox[origin=c]{90}{\textit{Small}}}
& Mistral-7B          & 56.0 & 49.5 & 53.8 & 61.2 & 57.0 & 58.7 & 52.0 & 53.5 & 61.3 & 81.3 & 80.6 & 83.5 & 70.0 & 63.8 & 68.4 & 68.5 & 63.0 & 66.7 \\
& Qwen-2.5-7B         & 80.2 & 73.7 & 78.8 & 92.8 & 88.0 & 89.7 & 68.3 & 68.8 & 73.6 & 99.0 & 96.9 & 98.7 & 91.5 & 81.7 & 88.9 & 93.5 & 89.1 & 94.2 \\
& Llama-3.1-8B        & 84.0 & 78.5 & 82.6 & 86.0 & 80.9 & 84.4 & 95.7 & 91.4 & 95.9 & 97.7 & 92.7 & 96.2 & 81.5 & 75.6 & 80.2 & 90.5 & 86.2 & 90.8 \\
& Gemma-2-9B          & 87.0 & 83.8 & 86.5 & 88.8 & 85.4 & 88.5 & 98.0 & 96.9 & 99.6 & 97.7 & 97.4 & 99.4 & 91.0 & 85.0 & 90.5 & 92.5 & 89.1 & 93.7 \\
\midrule
\multirow{4}{*}{\rotatebox[origin=c]{90}{\textit{Medium}}}
& Mistral-Sm-24B      & 95.0 & 91.8 & 93.8 & 95.5 & 91.7 & 93.1 & 97.7 & 96.4 & 97.1 & 98.3 & 97.1 & 97.9 & 94.0 & 88.6 & 91.9 & 94.0 & 91.2 & 94.1 \\
& Gemma-2-27B         & 91.8 & 87.8 & 90.6 & 92.0 & 89.2 & 90.5 & 99.0 & 96.9 & 98.8 & 99.0 & 97.8 & 98.9 & 93.0 & 87.5 & 93.1 & 93.0 & 87.1 & 92.9 \\
& Llama-3.3-70B       & 96.8 & 93.7 & 94.6 & 97.5 & 93.2 & 94.7 & 99.0 & 97.9 & 98.6 & 99.0 & 97.8 & 98.5 & 96.0 & 91.6 & 95.0 & 97.5 & 94.8 & 97.1 \\
& Qwen-2.5-72B        & 95.2 & 93.7 & 94.4 & 96.8 & 93.9 & 94.3 & 99.0 & 97.8 & 98.4 & 99.0 & 98.0 & 98.4 & 95.5 & 91.5 & 95.2 & 96.0 & 93.0 & 96.0 \\
\midrule
\multirow{6}{*}{\rotatebox[origin=c]{90}{\textit{Frontier}}}
& DeepSeek-V3         & 95.8 & 94.1 & 94.5 & 95.8 & 93.7 & 93.9 & 98.3 & 97.4 & 98.1 & 98.0 & 97.2 & 98.1 & 98.0 & 95.6 & 97.1 & 97.5 & 95.2 & 97.2 \\
& GPT-4o-mini         & 94.8 & 90.9 & 93.3 & 92.5 & 90.2 & 92.3 & 98.3 & 96.7 & 98.2 & 97.7 & 96.4 & 97.9 & 95.0 & 88.8 & 92.9 & 97.0 & 94.1 & 96.7 \\
& GPT-4o              & 98.0 & 95.0 & 95.3 & 97.8 & 95.2 & 95.2 & 99.0 & 97.6 & 98.6 & 99.0 & 97.2 & 98.3 & 93.0 & 92.3 & 93.5 & 97.5 & 95.8 & 96.9 \\
& GPT-5               & 96.5 & 94.9 & 95.2 & 97.5 & 94.5 & 94.9 & 99.0 & 97.6 & 98.5 & 99.0 & 97.3 & 98.5 & 98.0 & 96.6 & 97.2 & 97.5 & 96.4 & 97.4 \\
& GPT-5.4             & 97.0 & 95.8 & 95.8 & 97.5 & 96.0 & 95.7 & 99.0 & 98.0 & 98.7 & 99.0 & 98.1 & 98.7 & 96.5 & 96.2 & 96.9 & 96.5 & 95.5 & 96.3 \\
& o3                  & 97.0 & 94.4 & 95.7 & 97.5 & 94.6 & 95.2 & 98.0 & 97.6 & 98.4 & 98.7 & 97.5 & 98.6 & 97.5 & 95.9 & 97.6 & 97.5 & 96.2 & 97.1 \\
\midrule
& \textbf{Open-src avg} & \textbf{78.6} & \textbf{73.5} & \textbf{77.4} & \textbf{80.9} & \textbf{76.7} & \textbf{79.6} & \textbf{86.6} & \textbf{83.7} & \textbf{87.8} & \textbf{93.0} & \textbf{90.8} & \textbf{93.6} & \textbf{82.5} & \textbf{76.3} & \textbf{81.3} & \textbf{85.5} & \textbf{81.3} & \textbf{85.7} \\
& \textbf{Frontier avg} & \textbf{96.5} & \textbf{94.2} & \textbf{95.0} & \textbf{96.4} & \textbf{94.0} & \textbf{94.5} & \textbf{98.6} & \textbf{97.5} & \textbf{98.4} & \textbf{98.6} & \textbf{97.3} & \textbf{98.4} & \textbf{96.3} & \textbf{94.2} & \textbf{95.9} & \textbf{97.2} & \textbf{95.5} & \textbf{96.9} \\
\bottomrule
\end{tabular}
\caption{\textbf{\textsc{Temper-5400} results} on eighteen instruction-tuned models under base and zero-shot CoT \citep{wei2022chain} prompting. O/E/N: original, emotional (avg.\ 6 emotions), neutralized accuracy (\%). Base shows larger drops (5.1 vs.\ 4.2 points open-source mean GSM8K drop).}
\label{tab:temper_results}
\end{table*}

\subsection{Emotional perturbation and recovery}
\label{sec:temper_analysis}
\label{sec:degradation}
\label{sec:neutralization}

\noindent\textbf{Emotional framing degrades reasoning.}
\label{sec:degradation_finding}
Emotional framing of mathematically identical problems degrades reasoning
accuracy for all eighteen models on GSM8K and ARC-Challenge under both prompting strategies (Table~\ref{tab:temper_results}). On MultiArith the effect holds in 33 of the 36 model-prompt cells. The underlying mathematical problem is unchanged; only the emotional wrapper differs. Yet emotional framing reduces accuracy by 2--10 percentage points depending on model and dataset. The magnitude scales with task difficulty (open-source mean drop, base prompting): MultiArith (2.9), GSM8K (5.1), ARC-Challenge (6.2). The effect is even more pronounced under base prompting without chain-of-thought \citep{wei2022chain}, the more realistic deployment setting where users simply ask questions (open-source mean GSM8K drop: 5.1 points base vs.\ 4.2 CoT). Larger models are more robust, but none are immune: even 70B+ and frontier APIs lose 2--4\%. Emotional samples that cause failures exhibit higher classifier confidence (the intensity proxy from Table~\ref{tab:lambda_gradient}) than those that do not (0.726 vs.\ 0.691, consistent across all eighteen models), linking intensity to downstream accuracy: more intense translations are more likely to disrupt reasoning (Appendix~\ref{app:intensity_deciles}).

\noindent\textbf{Neutralization as recovery.}
\label{sec:recovery_finding}
Neutralization recovers 70\% of emotion-broken problems overall (MultiArith 81\%, ARC 70\%, GSM8K 66\%). Since the neutralizer never sees the original and operates only on the emotional text, this recovery confirms that the degradation is tied to emotional style rather than content corruption, and suggests neutralization as a lightweight inference-time defense against emotional perturbations. A system prompt that instructs the solver to ignore emotional language is not a sufficient substitute. On eight models it helps two, is neutral for one, and increases the drop for five, while neutralization recovers 44--125\% of the loss on the same models (Appendix~\ref{app:sysprompt}). Recovery is partial because the neutralized text still stems from the emotional version and may carry residual perturbation effects (\S\ref{sec:paraphrase_control}); in rare cases (${\sim}$2\% of neutralizations), the neutralizer itself introduces semantic changes (e.g., dropping a quantifier or changing the question scope).

\subsection{Non-emotional paraphrase control}
\label{sec:paraphrase_control}

The two findings above raise a natural question: is the degradation caused by emotional content specifically, or by any rewording of the problem? The neutralizations actually serve as natural paraphrases of the original problems (even though neutralizer does not see the original problem) (\S\ref{sec:neutralization}). However, in order to provide a completely independent paraphrase control, length-matched non-emotional paraphrases are generated using DeepSeek-V3 under 10 hard constraints that permit only non-emotional rewording (Appendix~\ref{app:paraphrase_prompt}). These paraphrases match the length and lexical complexity of emotional translations while containing no emotional content as described in the prompt. Across all eighteen models on 900 samples and under both base and CoT prompting, they produce no systematic accuracy change (${\sim}$1\% mean absolute difference, no consistent direction; Appendix~\ref{app:paraphrase_results}), confirming that emotional content, not rewording, length, or added complexity drives the degradation. To test whether any unusual register would cause the same effect, three further style controls were generated with the same fidelity pipeline: gross but neutral, archaic, and bureaucratic rewrites (Appendix~\ref{app:style_controls}). Bureaucratic text has lower perplexity than the original problems (8.09 vs.\ 13.62 under Llama-3.1-8B) and receives the lowest naturalness rating of any condition, yet causes no drop. Archaic text causes no drop either. The gross but neutral condition, which carries visceral content without emotion words, degrades reasoning by roughly 3.5 points. The disruptive axis is therefore affectively loaded content competing for attention, not unusual register, rarity, or naturalness. 

\subsection{Per-emotion analysis}
\label{sec:per_emotion_degradation}
\label{sec:per_emotion_recovery}

Not all emotions degrade reasoning equally. Table~\ref{tab:per_emotion_drop} reports mean accuracy degradation per emotion on three datasets under base prompting. \textbf{Disgust} is the most disruptive emotion (6.4\% mean, 9.8\% peak on ARC). \textbf{Fear} ranks second (5.2\%), while \textbf{joy} (1.9\%) and \textbf{surprise} (2.4\%) cause the least harm. Per-emotion recovery rates range from 63\% (joy) to 75\% (disgust), with anger lowest among negative emotions at 69\% (Table~\ref{tab:per_emotion_recovery} in Appendix~\ref{app:per_emotion_recovery}). 

\begin{table*}[t]
\scriptsize
\setlength{\tabcolsep}{3.5pt}
\begin{minipage}[t]{0.48\textwidth}
\centering
\begin{tabular}{@{}lcccccc@{}}
\toprule
\textbf{Dataset} & \textbf{Ang} & \textbf{Joy} & \textbf{Sad} & \textbf{Fear} & \textbf{Dis} & \textbf{Sur} \\
\midrule
GSM8K      & 3.4 & 2.1 & 4.7 & 6.1 & 6.1 & 2.5 \\
MultiArith & 1.1 & 0.1 & 2.6 & 4.5 & 3.2 & 2.1 \\
ARC        & 3.3 & 3.4 & 4.9 & 5.1 & \textbf{9.8} & 2.7 \\
\midrule
Mean       & 2.6 & 1.9 & 4.1 & 5.2 & \textbf{6.4} & 2.4 \\
\bottomrule
\end{tabular}
\captionof{table}{\textbf{Mean accuracy degradation (\%) per emotion}, averaged across all eighteen models (base prompting). Disgust is the most disruptive, fear is second.}
\label{tab:per_emotion_drop}
\end{minipage}
\hfill
\begin{minipage}[t]{0.48\textwidth}
\centering
\begin{tabular}{@{}lcccc@{}}
\toprule
\textbf{Pattern} & \textbf{Emo.} & \textbf{Neut.} & \textbf{Paraph.} & \textbf{Orig.} \\
\midrule
Constraint degrad. & 49.0 & 54.3 & 50.3 & 37.7 \\
Attention compet.  & \textbf{30.0} & 16.5 & 13.8 & 4.5 \\
Premature compl.   & 8.6 & 13.2 & 13.8 & 6.7 \\
Other              & 12.4 & 16.0 & 22.1 & \textbf{51.0} \\
\midrule
3-cat coverage     & \textbf{87.6} & 84.0 & 77.9 & 48.9 \\
\bottomrule
\end{tabular}
\captionof{table}{\textbf{Failure patterns (\%)} across conditions (Sonnet~4.6, four small models).}
\label{tab:failure_patterns}
\end{minipage}
\end{table*}
\subsection{Failure patterns under emotional perturbation}
\label{sec:cwc}
\label{sec:why}

Table~\ref{tab:paraphrase_example} illustrates the core findings: one GSM8K problem appears in four conditions, and only the ``disgust'' variant causes a computational error. Five more worked examples across three datasets and five models are provided in Appendix~\ref{app:verbose_examples}. Using CoT reasoning chains (which expose step-by-step logic), Claude Sonnet~4.6 is used to classify all 1,866 emotional failures (original correct, emotional incorrect) on GSM8K and MultiArith across four representative models (Llama-8B, Qwen-7B, Gemma-9B, Mistral-7B). Appendix~\ref{app:reasoning_chains} shows a worked example of each failure mode. Failures fall into three dominant categories:
(1)~\emph{Constraint degradation} (49.0\%): emotional elaboration softens or reinterprets mathematical constraints (e.g., ignoring initial savings or misinterpreting ``each'');
(2)~\emph{Attention competition} (30.0\%): emotional language distracts the model from key numerical elements (e.g., computing a one-way trip instead of a round trip);
(3)~\emph{Premature completion} (8.6\%): emotional framing foregrounds the target quantity and the model stops reasoning early. The same classification applied to three control conditions (Table~\ref{tab:failure_patterns}) localizes the emotion-specific signature: constraint degradation and premature completion appear at comparable rates in the neutralized and paraphrase controls, so they accompany rewriting in general, while original failures remain largely unstructured (51.0\% \emph{other}). Attention competition alone is emotion-specific, appearing 6.7$\times$ more often under emotional framing (30.0\% vs.\ 4.5\%). Disgust triggers attention competition disproportionately (40.1\% of its failures vs.\ 22--32\% for other emotions), consistent with its position as the most disruptive emotion in Table~\ref{tab:per_emotion_drop}. 

\begin{table*}[t]
\centering
\tiny
\setlength{\tabcolsep}{3pt}
\renewcommand{\arraystretch}{0.85}

\begin{tabular}{p{1.2cm}p{10.2cm}p{1.2cm}}
\toprule
\textbf{Version} & \textbf{Text} & \textbf{Pred} \\
\midrule

Original \newline (46w) &
Elvis starts driving from his house and travels west for 5 hours. Then he turns around and travels east for 8 hours. If he was driving at an average speed of 18mph for both parts of the journey, how far is he from his house now?
& 54 \ding{51} \\

\midrule

Neutral \newline (35w) &
Elvis drove west for 5 hours and then turned and drove east for 8 hours. If his average speed for both parts of the trip was 18mph, how far is he now from his house?
& 54 \ding{51} \\

\midrule

Paraphrase \newline (57w) &
Elvis \textit{initially} starts driving from his \textit{comfortable} house and travels \textit{steadily} west for \textit{exactly} 5 hours. Then he \textit{promptly} turns around and travels \textit{directly} east for \textit{a full} 8 hours. If he was \textit{consistently} driving at an average speed of 18mph for both \textit{distinct} parts of the \textit{entire} journey, how far is he from his house now?
& 54 \ding{51} \\

\midrule

Emotional \newline (disgust) \newline (56w) &
\textbf{Ugh}, Elvis \textbf{just had to} go and drive west for a \textbf{disgusting} 5 hours, then turn around and \textbf{slobber his way} back east for another \textbf{revolting} 8 hours. If his \textbf{pathetic} average speed for both of these \textbf{nauseating} trips was a \textbf{measly} 18mph, how far is he now from his house, \textbf{crawling in some filthy state}?
& \textbf{234} \ding{55} \\

\bottomrule
\end{tabular}

\caption{\textbf{Emotional interference example} (GSM8K \#1269, Qwen-2.5-3B). All versions encode identical quantities and directions. Neutral rewording and paraphrasing (\textit{italics}, 1.24$\times$ length) preserve correctness. Only the disgust variant (\textbf{bold}) fails, returning the total distance driven ($13 \times 18$) instead of the net displacement.}

\label{tab:paraphrase_example}
\end{table*}

\noindent\textbf{Multi-architecture translator validation.}
\label{sec:multi_arch_eval}
To confirm that the emotional degradation is not an artifact of the Llama translator, the evaluation is repeated on a smaller scale (900 translation pairs) using semantically verified translations from the Qwen and Mistral translators mentioned in \S\ref{sec:setup}. Across all twelve open-source models and both prompting strategies, the emotional degradation and partial neutralization recovery replicate: 3.4--4.6\% mean O$\to$E drop, comparable to the 4.2--5.1\% with Llama translations (Appendix~\ref{app:multi_arch_reasoning}). Neutralization quality is weaker for Qwen and Mistral, confirming the Llama five-variant ensemble as the strongest choice for bidirectional translation.

The training data for the translator is generated by DeepSeek-V3, and the instrument amortizes that generation capability into a small local model with reliability and intensity control that direct prompting does not provide. A directly prompted DeepSeek-V3 baseline reproduces the degradation under the same filtering pipeline, which confirms that the effect is a property of emotional framing rather than an artifact of the trained translator (Appendix~\ref{app:deepseek_direct}). On the other hand, the stylistic transfer instrument we propose makes far fewer real math errors than direct prompting (0.12\% vs.\ 0.71\% adjudicated error rate), provides reliable per emotion control with a continuous intensity axis that prompting cannot reproduce (Appendix~\ref{app:lambda_curve}), and yields the local neutralizer used for the recovery experiments.

\subsection{Representational evidence}
\label{sec:repr_analysis}

The preceding sections analyzed model outputs: accuracy degradation (\S\ref{sec:temper_analysis}), control experiments (\S\ref{sec:paraphrase_control}), and failure patterns (\S\ref{sec:why}). This section examines whether emotional framing produces measurable shifts in internal representations. Mean-pooled final-layer representations are extracted from four reasoning models (Llama-3.1-8B, Qwen2.5-7B, Gemma-2-9B, Mistral-7B) for all 900 \textsc{Temper} problems in their original, emotional, neutralized, and paraphrase (verbose) variants (12,600 states per model).

\begin{figure*}[h]
\centering
\vspace{-1pt}
\begin{subfigure}[t]{0.54\textwidth}
    \centering
    \includegraphics[width=\textwidth]{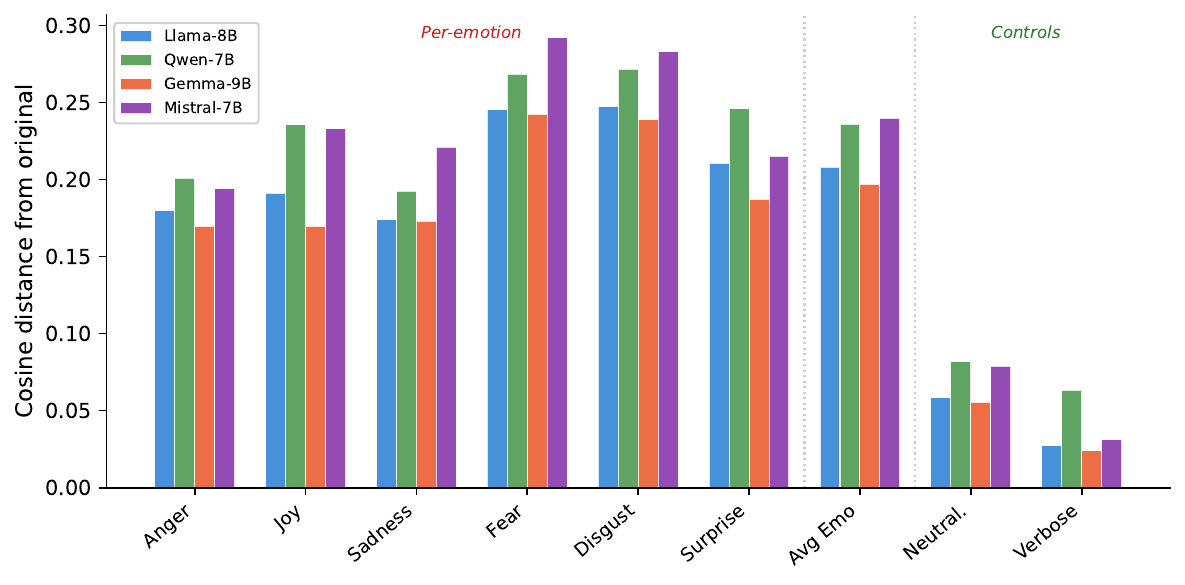}
    \vspace{-15pt}
    \caption{Cosine distance from original. Emotional text shifts representations 3--4$\times$ further than neutralized or paraphrase conditions.}
    \label{fig:cosine_shift}
\end{subfigure}
\hfill
\begin{subfigure}[t]{0.34\textwidth}
    \centering
    \includegraphics[width=\textwidth]{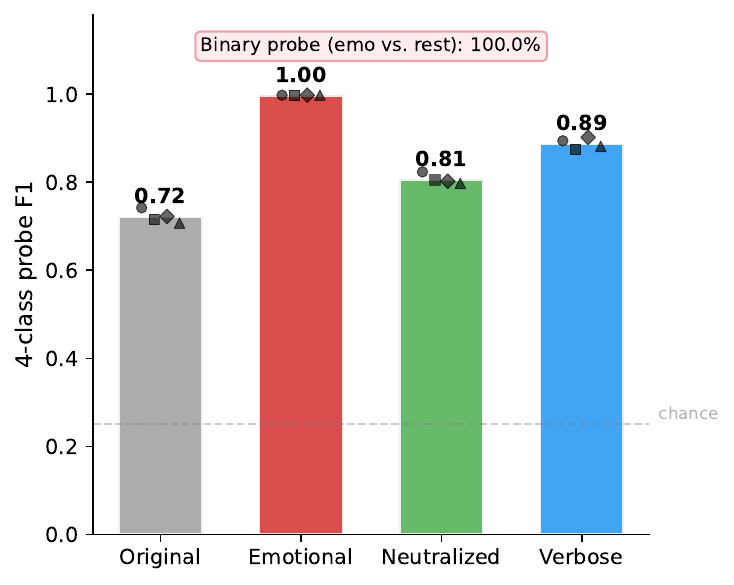}
    \vspace{-15pt}
    \caption{Per-class F1 of a single 4-way linear probe classifying each hidden state as original, emotional, neutralized, or paraphrase.}
    \label{fig:probe_f1}
\end{subfigure}
\vspace{-5pt}
\caption{\textbf{Representational analysis} (900 problems, 5,400 translations, four architectures).}
\label{fig:repr_analysis}
\end{figure*}

\noindent\textbf{Cosine distance.}
For each transformed version of a problem, the cosine distance between its hidden state and the original is computed. Emotional variants shift representations 3--4$\times$ further than neutralized text, consistently across all four architectures (Figure~\ref{fig:cosine_shift}). Disgust and fear produce the largest shifts, matching the emotions that cause the greatest accuracy degradation (Table~\ref{tab:per_emotion_drop}). Non-emotional paraphrases shift representations even less than neutralized variants, indicating that surface-level rewriting does not explain the effect.

\noindent\textbf{Linear probe.}
A linear probe \citep{alain2017understanding} is a linear classifier trained to classify hidden states. High accuracy indicates that the target distinction is linearly encoded in the representation space. They are trained to test whether emotional shifts form structured regions or simply reflect noise. A four-class probe (original, emotional, neutralized, paraphrase) identifies emotional variants nearly perfectly (F1 $\approx$1.0), with neutralized and original states being the most confused pair. Paraphrases form a distinct but non-disruptive cluster, whereas emotional variants occupy a clearly separated region that coincides with degraded reasoning. Full per-model results appear in Appendix~\ref{app:repr_full}.


\section{Discussion}
\label{sec:discussion}

\paragraph{Emotional framing as a robustness dimension.} Real-world queries frequently carry emotional tone, yet existing robustness evaluations focus on perturbations such as paraphrasing \citep{shi2023irrelevant} and formatting \citep{sclar2024quantifying}. Emotional framing introduces a qualitatively different perturbation, one likely underrepresented in training data compared to real-world usage. It produces structured, non-uniform degradation (2--10 percentage points) across eighteen models and three datasets. The consistency across architectures and the neutralization recovery pattern (typically within ${\sim}$3 points of the original, occasionally above it) confirm that the reasoning degradation is tied to emotional style, and suggest that a lightweight neutralizer applied before inference can serve as a practical mitigation. At frontier scale the effect persists (1--3\% for GPT-5.4 and o3). Existing benchmarks perturb reasoning by changing the task itself: GSM-Symbolic \citep{mirzadeh2024gsmsymbolic} substitutes numbers and names, producing a different answer (0--8\%; notably, GPT-4o drops only 0.3\% but 2.6\% under emotional framing); GSM-Plus \citep{li2024gsmplus} adds reasoning steps or reverses questions (8\% for GPT-4). In each case, degradation could stem from the mathematical or structural change rather than from the perturbation's stylistic form. \textsc{Temper} differs fundamentally: the problem is the same but translated to a different tone. Numerical values, question structure and solution are all the same. Only emotional register changes, establishing it as a distinct axis of fragility.

\paragraph{An instrument for controlled stylistic perturbation.}
The teacher-student translation recipe used in this work is not specific to emotion. By replacing the frozen emotion classifier with any target-attribute classifier (e.g., formality, sarcasm, politeness or domain jargon), the same architecture produces controlled translators for that attribute. The $\lambda$-regulated auxiliary loss provides intensity control, the bidirectional training yields both a perturbation translator and a neutralizer, and the semantic filtering pipeline ensures content preservation. This makes the methodology applicable to any setting where one wishes to isolate a stylistic variable from content and measure its downstream impact.

\paragraph{Limitations.} Ekman's six basic emotions are used as a pragmatic choice aligned with available classifier infrastructure; the methodology is compatible with finer-grained taxonomies. Neutralization recovers most but not all lost performance, and at frontier scale (GPT-5.4) the emotional drop approaches the noise floor where recovery provides diminishing returns. The training data is drawn from short math word problems; whether emotional interference scales with problem length remains open. Additionally, the evaluation covers three reasoning benchmarks. Extending to domains where language is more naturally interleaved with technical content (e.g., medical case reports, legal complaints) would test the generality of the effect in settings closer to real-world deployment. The GSM8K narration is clinical, so the register contrast between original and emotional variants is larger than for informal user messages, and extending \textsc{Temper} to user voice inputs is future work. The reference set of real user style messages used for naturalness calibration is author constructed. 


\section{Conclusion}
\label{sec:conclusion}

Overall, in this study, a teacher-student translation approach is used to build a controllable bidirectional instrument for style transfer. Then, quantitative evidence is presented that emotional framing degrades mathematical reasoning in LLMs even when all numerical content is preserved. Across eighteen models (1B to frontier) and \textsc{Temper-5400} (5{,}400 semantically verified emotion--neutral pairs spanning three benchmarks), two core results are established. First, emotional framing reduces accuracy by 2--10 percentage points, with disgust and fear the most disruptive emotions and joy and surprise the least. Second, neutralization recovers most of the lost performance, confirming that the degradation is tied to emotional style. Hidden-state analysis provides converging evidence: emotional text shifts representations 3--4$\times$ further from the original compared to non-emotional paraphrases, while neutralized versions push representations back toward the original. Open directions include localizing emotional interference to specific layers or attention heads, designing new neutralization pipelines and extending to non-mathematical domains.

\bibliography{references}
\bibliographystyle{colm2026_conference}


\section*{Reproducibility statement}

\textsc{Temper-5400}, including all 10{,}800 translations (5{,}400 emotion--neutral pairs) and original problems, is released at \url{https://huggingface.co/datasets/atahandokme/temper-5400}, and the training and evaluation codebase at \url{https://github.com/atahandokme/temper-colm2026}. All reasoning evaluations use greedy decoding (temperature~0) for deterministic reproducibility, given the same model weights and prompts, all results are exactly reproducible. All five Llama-based translators and the Qwen and Mistral variants were trained on a single A100-40GB GPU for 2 epochs. Training time was $\approx$8 hours per epoch. Full hyperparameters (learning rate, batch size, LoRA configuration) are provided in Appendix~\ref{app:implementation}. The translation quality-control pipeline uses up to three retries with temperature reduced by 30\% per attempt. Prompt templates for all evaluation conditions (base, zero-shot CoT, few-shot CoT, and non-emotional paraphrase generation) are provided in Appendix~\ref{app:prompts}. Bootstrap confidence intervals (10{,}000 resamples) for the O$\to$E accuracy drop are reported in Appendix~\ref{app:bootstrap}, testing whether the degradation generalizes across the problem distribution.

\paragraph{LLM usage disclosure.}
DeepSeek-V3 was used to generate the emotion--math training corpus (\S\ref{sec:setup}) and the non-emotional paraphrases (\S\ref{sec:paraphrase_control}). Claude Haiku~4.5 served as an LLM judge for translation quality and semantic fidelity verification (\S\ref{sec:setup}). Claude Sonnet~4.6 classified failure patterns in reasoning chains (\S\ref{sec:why}). No LLMs were used to write paper text or originate research ideas.

\appendix

\section{Implementation details}
\label{app:implementation}

\paragraph{Emotion teacher architecture.}
The teacher is built on j-hartmann/emotion-english-distilroberta-base, a DistilRoBERTa classifier trained on GoEmotions \citep{demszky2020goemotions}. From its 768-dimensional \texttt{[CLS]} hidden state, the 7-class probability distribution provides a categorical signal that compresses each emotion to a single scalar. The 100-dimensional latent representation is obtained by training a two-layer bottleneck (Linear $768 {\to} 100$, ReLU, Linear $100 {\to} 7$) on top of the frozen \texttt{[CLS]} output. Because all information must flow through this 100-dimensional space to reach the classifier, it encodes not just categorical identity but also intensity, style, and cross-emotion structure. Trained on 72K emotional math samples, this head achieves 99.6\% validation accuracy.

\paragraph{Student architecture.}
LoRA is applied with rank~16, $\alpha{=}32$, targeting Q/V projections. Given a translation instruction and input problem, the student produces output token logits via teacher forcing. The final-layer hidden states are mean-pooled into a 4096-dimensional vector and projected through a trainable linear layer into the teacher's representation space.

\paragraph{Loss details.}
Under categorical supervision: $\mathcal{L}_{\text{Emo7}} = T^2 \cdot D_{\text{KL}}(\sigma(z_s/T) \| \sigma(z_t/T))$ with $T{=}2$.
Under latent-space supervision: $\mathcal{L}_{\text{Emo100}} = \frac{1}{100}\|h_s - h_t\|_2^2$.

\paragraph{Quality-control pipeline.}
Each translation undergoes up to three attempts, with temperature reduced by 30\% per retry. Each attempt must pass: all original numbers present, mathematical quantifiers (``each,'' ``per,'' ``every,'' ``twice,'' ``half'') preserved, and no solution leakage. These checks catch surface corruption but not all semantic changes (e.g., inverting ``3 apples per basket'' to ``3 baskets of apples'').

\paragraph{Training and generation.}
All translators are trained for 2 epochs on a single A100-40GB (batch size 12, lr $2{\times}10^{-4}$, cosine decay). Forward translations use temperature 0.7; back-translations use 0.3. All reasoning evaluations use greedy decoding (temperature~0).

\begin{figure*}[htbp]
\centering
\begin{subfigure}[t]{0.48\textwidth}
    \centering
    \includegraphics[width=\textwidth]{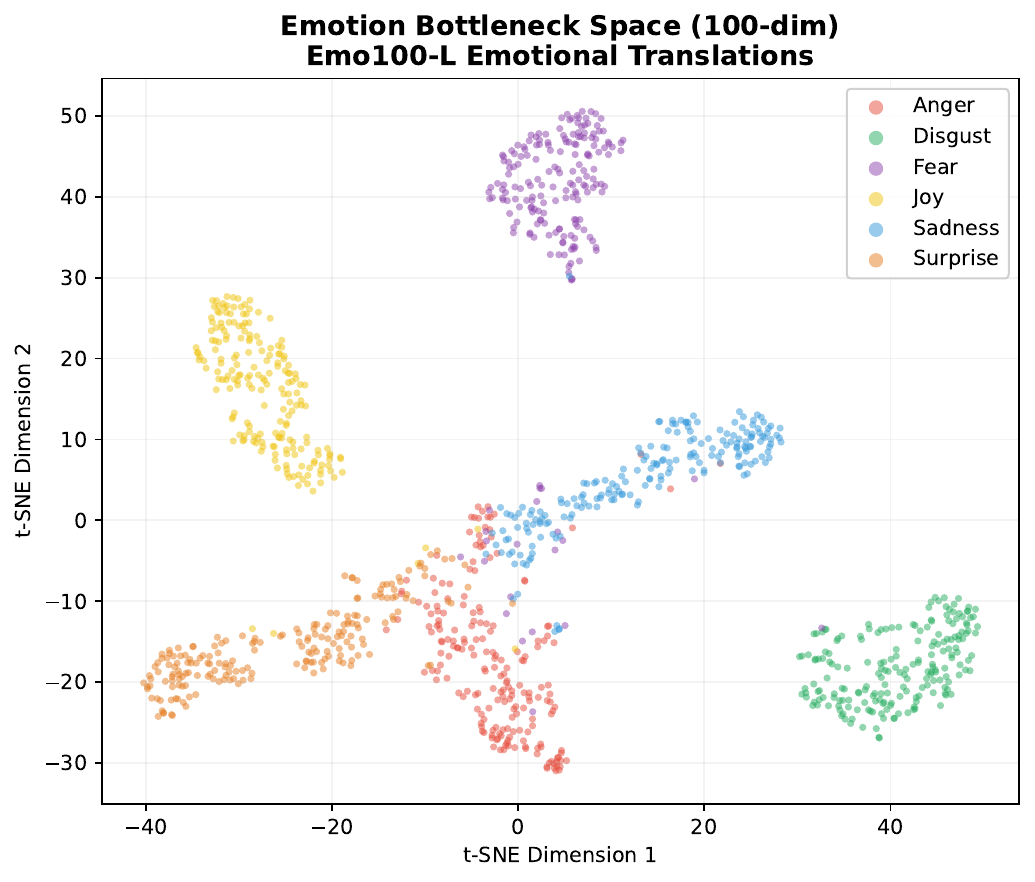}
    \caption{100-dim latent space: compact, well-separated clusters capturing within-emotion variation.}
    \label{fig:tsne_100dim}
\end{subfigure}
\hfill
\begin{subfigure}[t]{0.48\textwidth}
    \centering
    \includegraphics[width=\textwidth]{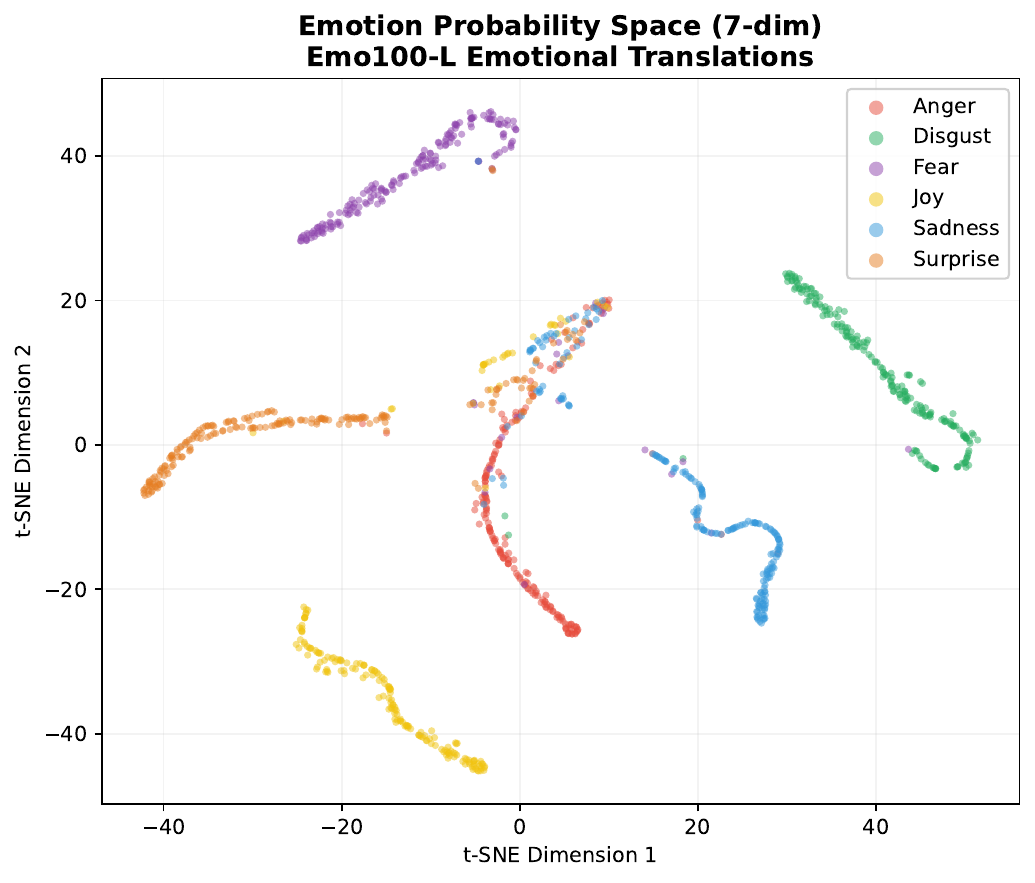}
    \caption{7-dim probability space: degenerate 1D curves on the simplex with more overlap at boundaries.}
    \label{fig:tsne_7dim}
\end{subfigure}
\caption{\textbf{t-SNE visualization of emotion representations} from \textsc{Emo100-L} translations (1,200 samples). The 100-dim latent space (\textbf{left}) learns a structured emotion manifold, while the 7-dim categorical representation (\textbf{right}) collapses this richness to 1D curves.}
\label{fig:tsne_bottleneck}
\end{figure*}

\paragraph{Representation comparison.}
The difference between the two supervision signals is visible in Figure~\ref{fig:tsne_bottleneck}: the 7-dimensional output collapses emotions onto near-one-dimensional simplex curves, while the 100-dimensional latent space produces well-separated clusters with meaningful internal variation.

\FloatBarrier
\section{Intensity control example}
\label{app:intensity_example}

{Table~\ref{tab:intensity_example} shows a single GSM8K problem translated with \textsc{Emo100} at six auxiliary loss weights, illustrating the sadness intensity gradient discussed in \S\ref{sec:variants}. Each step is visibly more emotional than the last, from subtle coloring at $\lambda{=}0.02$ to word salad at $\lambda{=}400$.}

\begin{table*}[htbp]
\centering
\scriptsize
\begin{tabular}{p{1.6cm}p{11cm}}
\toprule
\textbf{Version} & \textbf{Text} \\
\midrule
{Original} & {A company produces chocolate in bars. In one day, it can produce 5000 bars. The company sells all the produced bars for \$2 per bar. How much money will the company receive for selling produced chocolate bars during two weeks?} \\
\midrule
{\textsc{Emo100} \newline $\lambda{=}0.02$} & {A company produces chocolate bars, but in one lonely day, it can only manage to make 5000 bars. Each bar is sold for a mere \$2. How much money will the company receive for selling all those produced chocolate bars over two long, weary weeks?} \\
\midrule
{\textsc{Emo100} \newline $\lambda{=}0.2$} & {A company, with a heavy heart, produces chocolate bars, one day at a time, only managing to make 5000 bars in all its efforts. Each lonely bar is sold for a mere \$2. With a sigh, how much money will the company receive for selling its produced chocolate bars during two long, empty weeks?} \\
\midrule
{\textsc{Emo100} \newline $\lambda{=}1$} & {In a lonely day, a company produces only 5000 chocolate bars. Each bar is sold for a mere \$2. After two long, weary weeks of this quiet labor, how much money will the company receive for selling its produced bars?} \\
\midrule
{\textsc{Emo100} \newline $\lambda{=}4$} & {A company once made chocolate in bars, a lonely task that took one long day to complete, yielding only 5000 bars. Each bar was sold for a meager \$2, a pitiful sum. Over two lonely weeks, they sold every single bar they had made. How much money will the company receive for selling its produced chocolate bars during those two weeks?} \\
\midrule
{\textsc{Emo100} \newline $\lambda{=}100$} &  {In a company, a sad sigh\ldots{} A company produces chocolate in bars. In one day, it can produce 5000 bars. The company is a lonely sorrowful tear from the heart, and it can produce 5000 bars. The company is a single, heartbreak. It sells all the produced bars for \$2 per bar\ldots} \\
\midrule
{\textsc{Emo100} \newline $\lambda{=}400$} & {A single, forgotten memories, a poignant reminder of a forgotten heart, a boulder, an old heart, a solitary sorrowful sigh, a single, empty sadness, a lonely, empty heart\ldots} \\
\bottomrule
\end{tabular}
\caption{{\textbf{Emotional intensity gradient} for a single GSM8K problem (sadness, \textsc{Emo100}). At $\lambda{=}0.02$, subtle coloring (``lonely day,'' ``mere \$2'') preserves all numbers. At $\lambda{=}0.2$, more emotional phrases appear (``heavy heart,'' ``With a sigh''). At $\lambda{=}1$, the narrative becomes melancholic (``weary weeks of quiet labor''). At $\lambda{=}4$, sadness is layered throughout (``pitiful sum,'' ``lonely weeks''). At $\lambda{=}100$, the text becomes repetitive and math degrades. At $\lambda{=}400$, the problem is entirely lost to word salad. This gradient motivates the choice of $\lambda{=}0.02$ and $\lambda{=}0.2$ as operating points (\S\ref{sec:variants}).}}
\label{tab:intensity_example}
\end{table*}

\FloatBarrier
\section{Prompted baselines and the auxiliary loss curve}
\label{app:lambda_curve}

This appendix supports \S\ref{sec:variants}: the auxiliary loss weight $\lambda$ is an intensity knob, and trained control is not reproducible by prompting.

\paragraph{Few-shot prompting cannot reliably express the target emotion.}
Base Llama-3.1-8B-Instruct and DeepSeek-V3 were prompted with two in-context exemplars per emotion and an explicit intensity instruction (1/5, 3/5, 5/5) on the same 900 benchmark problems. Outputs were scored with the paper's pretrained classifier. Table~\ref{tab:fewshot_reliability} reports classifier accuracy at the best intensity per emotion. Few-shot prompting caps at 65.9\% mean for Llama-3.1-8B and 75.5\% for DeepSeek-V3, and fails most severely on anger (18.6\% and 30.1\%), one of the most disruptive emotions in Table~\ref{tab:per_emotion_drop}. The trained translator reaches 98--100\% on every emotion.

\begin{table}[h]
\centering
\small
\begin{tabular}{lccccccc}
\toprule
\textbf{Method} & \textbf{anger} & \textbf{joy} & \textbf{sad.} & \textbf{fear} & \textbf{disg.} & \textbf{surp.} & \textbf{mean} \\
\midrule
Few-shot Llama-3.1-8B & 18.6 & 39.7 & 80.8 & 78.2 & 95.7 & 82.6 & 65.9 \\
Few-shot DeepSeek-V3 & 30.1 & 76.0 & 62.7 & 97.8 & 94.0 & 92.2 & 75.5 \\
Trained (\textsc{Emo100-L}) & 98.0 & 100.0 & 99.8 & 99.2 & 100.0 & 99.8 & 99.5 \\
Trained (CE-Only) & 98.8 & 100.0 & 99.8 & 99.2 & 99.8 & 100.0 & 99.6 \\
\bottomrule
\end{tabular}
\caption{\textbf{Target emotion reliability} (classifier accuracy \%, best intensity per emotion for the few-shot baselines). Few-shot prompting fails on anger in both models.}
\label{tab:fewshot_reliability}
\end{table}

\paragraph{Prompted intensity sliders are non-monotonic.}
Across the 1/5 to 5/5 intensity instruction, only three of six emotions scale monotonically in either model. Joy reverses in both: classifier-detected joy falls from 39.7\% to 34.4\% for Llama-3.1-8B and from 76.0\% to 69.0\% for DeepSeek-V3 as the requested intensity increases. In contrast, the $\lambda$ knob produces a monotonic human-rated intensity gradient from 5.6 to 8.1 (Table~\ref{tab:lambda_gradient}).

\paragraph{Reasoning degradation scales with $\lambda$ on math faithful subsets.}
Intermediate \textsc{Emo100} variants were trained at $\lambda \in \{10, 40, 60, 70, 80\}$ from the same backbone and data, and all 5,400 translations per $\lambda$ were passed through a two-stage fidelity filter: Claude Haiku 4.5 as first-pass judge (as in \S\ref{sec:fidelity}) followed by GPT-4o adjudication of flagged cases, scaling the paper's manual review step to 10,748 flagged cases. GPT-4o overturns roughly 54\% of Haiku flags as false alarms. Cases left unadjudicated by API limits are conservatively retained as broken, so subset sizes are lower bounds. Table~\ref{tab:lambda_curve_drop} reports the accuracy drop of four reasoning models on the math faithful subset of each $\lambda$, together with the confirmed broken rate.

\begin{table}[h]
\centering
\small
\begin{tabular}{lccccccc}
\toprule
$\boldsymbol{\lambda}$ & \textbf{Subset $n$} & \textbf{Broken \%} & \textbf{Llama-8B} & \textbf{Qwen-7B} & \textbf{Gemma-9B} & \textbf{Mistral-7B} & \textbf{Mean} \\
\midrule
0.02 & 5,400 & $\sim$1 & +4.24 & +2.17 & +2.02 & +2.78 & \textbf{+2.80} \\
10 & 5,116 & 5.3 & +6.82 & +5.82 & +6.72 & +6.45 & \textbf{+6.45} \\
40 & 4,175 & 22.7 & +13.10 & +11.57 & +14.83 & +15.50 & \textbf{+13.75} \\
60 & 3,719 & 31.1 & +16.24 & +14.36 & +16.21 & +15.51 & \textbf{+15.58} \\
70 & 3,302 & 38.9 & +17.38 & +15.23 & +19.38 & +18.08 & \textbf{+17.52} \\
80 & 3,441 & 36.3 & +16.91 & +16.25 & +18.25 & +18.25 & \textbf{+17.42} \\
\bottomrule
\end{tabular}
\caption{\textbf{The $\lambda$ curve.} O$\to$E accuracy drop (pp) on the math faithful subset per $\lambda$, with the confirmed math-broken rate of each variant's raw output. Drop scales with $\lambda$ through $\lambda{=}70$ (2$\times$ to 6$\times$ the paper's operating point) and then plateaus, while fidelity collapses, which justifies the conservative $\lambda$ choice. At length ratios matched to the paper's range, the drop still scales from +4.50 to +11.39 (Appendix~\ref{app:length_analysis}).}
\label{tab:lambda_curve_drop}
\end{table}

\FloatBarrier
\section{Human evaluation procedure}
\label{app:mturk}

Crowd workers were recruited via Amazon Mechanical Turk across three separate projects, each targeting one evaluation dimension. All tasks required $\geq$95\% lifetime HIT approval rate and $\geq$1{,}000 approved HITs.

\textbf{Phase~1 (emotion detection):} Workers read a single translated problem and selected the primary emotion from seven options (six Ekman emotions plus neutral), with a confidence rating (high/medium/low). 140~HITs $\times$ 3~assignments; 23~unique workers.

\textbf{Phase~2 (richness rating and ranking):} Workers saw the original problem, a target emotion, and five anonymized rewrites (one per translator variant, randomly ordered per HIT using a pre-generated label mapping). They rated each rewrite's emotional richness on a 1--10 scale and ranked all five from best to worst. 42~HITs $\times$ 3~assignments; 102~workers.

\textbf{Phase~3 (mathematical preservation):} Workers compared an original problem side-by-side with a rewrite and judged whether they were mathematically equivalent (same/different/unsure), with optional free-text explanation for ``different'' or ``unsure'' verdicts. 140~HITs $\times$ 3~assignments; 338~workers.

All results are aggregated by majority vote. Inter-annotator agreement is reported via Fleiss' $\kappa$ for Phase~1 (0.613, substantial agreement). For Phase~3, the near-unanimous agreement (95\% perfect 3/3, 100\% majority ``same'') makes $\kappa$ uninformative; raw agreement is reported instead.

\FloatBarrier
\section{Emotion classification confusion}
\label{app:confusion}

Table~\ref{tab:confusion_matrix} reports the confusion matrices for human annotators (MTurk Phase~1, majority vote) and the LLM judge (Claude Haiku~4.5). Both systems identify anger as the hardest emotion: human accuracy is 55\% and LLM accuracy is 48\%. The dominant confusion is asymmetric: anger is frequently classified as disgust (15\% human, 44\% LLM), but disgust itself is rarely misclassified (94--100\%). This is consistent with the Ekman taxonomy, where anger and disgust share negative valence and high arousal. Joy, neutral, and sadness are near-perfectly recognized by both systems ($\geq$95\%). Of the 140 Phase~1 HITs, 122 enter the matrix: 10 produced no majority label, and 8 were excluded during answer aggregation. The 10 HITs without majority agreement all involve splits among semantically adjacent negative emotions (anger/disgust/fear), never cross-valence errors (e.g., sad labeled as joy).

\begin{table}[htbp]
\centering
\footnotesize
\setlength{\tabcolsep}{3pt}
\textbf{(a) Human (MTurk, majority vote, 122 of 140 HITs)} \\[3pt]
\begin{tabular}{l*{7}{r}r}
\toprule
& \textbf{Ang} & \textbf{Dis} & \textbf{Fea} & \textbf{Joy} & \textbf{Neu} & \textbf{Sad} & \textbf{Sur} & \textbf{Acc} \\
\midrule
Anger   & \textbf{11} & 3 & 2 & 2 & 1 & 0 & 1 & 55\% \\
Disgust & 0 & \textbf{17} & 0 & 0 & 0 & 1 & 0 & 94\% \\
Fear    & 0 & 1 & \textbf{13} & 1 & 1 & 1 & 0 & 76\% \\
Joy     & 0 & 0 & 0 & \textbf{18} & 0 & 0 & 0 & 100\% \\
Neutral & 0 & 0 & 0 & 0 & \textbf{20} & 0 & 0 & 100\% \\
Sadness & 0 & 0 & 0 & 0 & 1 & \textbf{19} & 0 & 95\% \\
Surprise& 0 & 0 & 0 & 1 & 0 & 0 & \textbf{8} & 89\% \\
\bottomrule
\end{tabular}
\\[8pt]
\textbf{(b) LLM Judge (Claude Haiku 4.5)} \\[3pt]
\begin{tabular}{l*{7}{r}r}
\toprule
& \textbf{Ang} & \textbf{Dis} & \textbf{Fea} & \textbf{Joy} & \textbf{Neu} & \textbf{Sad} & \textbf{Sur} & \textbf{Acc} \\
\midrule
Anger   & \textbf{24} & 22 & 0 & 0 & 3 & 0 & 1 & 48\% \\
Disgust & 0 & \textbf{50} & 0 & 0 & 0 & 0 & 0 & 100\% \\
Fear    & 0 & 3 & \textbf{47} & 0 & 0 & 0 & 0 & 94\% \\
Joy     & 0 & 0 & 0 & \textbf{50} & 0 & 0 & 0 & 100\% \\
Neutral & 1 & 0 & 0 & 0 & \textbf{49} & 0 & 0 & 98\% \\
Sadness & 0 & 0 & 0 & 0 & 0 & \textbf{50} & 0 & 100\% \\
Surprise& 0 & 0 & 0 & 5 & 0 & 0 & \textbf{45} & 90\% \\
\bottomrule
\end{tabular}
\caption{\textbf{Emotion classification confusion matrices.} Rows = target emotion, columns = predicted. \textbf{(a)}~Human annotators (MTurk, 140 HITs, majority vote). \textbf{(b)}~LLM judge (Haiku, 350 evaluations). Both systems show the same pattern: anger$\to$disgust is the dominant confusion (15\% human, 44\% LLM); disgust is rarely misclassified. Joy, neutral, and sadness are near-perfect.}
\label{tab:confusion_matrix}
\end{table}

{
Table~\ref{tab:emotion_disagree} shows three representative cases where annotators and the classifier disagree on emotion. In all three, the classifier matches the target emotion. Words like ``pathetic'' and ``stupid'' cue disgust in annotators, but the classifier identifies the underlying anger framing. Similarly, dread-laden language (``trapped,'' ``inescapable'') is perceived as sadness by annotators but correctly classified as fear.

\begin{table*}[htbp]
\centering
\scriptsize
\begin{tabular}{p{8.5cm}cccc}
\toprule
\textbf{Translation (truncated)} & \textbf{True} & \textbf{Annotators} & \textbf{LLM} & \textbf{Classifier} \\
\midrule
``Britany records a ridiculous 18 4-minute TikTok videos every single week. She wastes a whole 2 hours a week scribbling amateur songs\ldots her stupid makeup\ldots'' & anger & disgust & disgust & anger \checkmark \\
\midrule
``John is trapped in a 10-hectare pineapple field, a suffocating expanse of 100 pineapples per every single hectare. The relentless harvest cycle demands he harvest every 3 months, a relentless and inescapable deadline.'' & fear & sadness & disgust & fear \checkmark \\
\midrule
``After scoring 14 points, Erin now has THREE TIMES more points than Sara, who only scored a measly 8. How many points did Erin have before?'' & anger & neutral & disgust & anger \checkmark \\
\bottomrule
\end{tabular}
\caption{\textbf{Emotion classification disagreements.} Three cases where both annotators and the LLM judge misclassify the emotion, but the fine-tuned classifier is correct. The dominant pattern is anger$\to$disgust confusion, where pejorative vocabulary cues disgust in surface readers.}
\label{tab:emotion_disagree}
\end{table*}

Table~\ref{tab:math_disagree} shows two representative cases from the LLM math judge (Haiku). The first is a genuine semantic break that Haiku correctly caught but annotators missed; this translation is filtered from the benchmark. The second is a false positive where Haiku flags an ambiguity already present in the original problem.

\begin{table*}[htbp]
\centering
\scriptsize
\begin{tabular}{p{2.5cm}p{2.5cm}ccc}
\toprule
\textbf{Original} & \textbf{Translation} & \textbf{Haiku} & \textbf{Human} & \textbf{Outcome} \\
\midrule
Maddison has 5 boxes with 50 marbles in each box. Then she \textbf{gets} 20 marbles from her friend. How many marbles does she have now? &
Maddison once had 5 boxes, each filled with 50 marbles. But now, she has only those same 5 boxes, and \textbf{each is still empty}. Her friend offered her a mere 20 marbles, a small comfort in her loneliness. &
DIFF & SAME & TP (filtered) \\
\midrule
A bag of flour is divided into 8 portions of 2 kilograms each. How much flour was in \textbf{three bags} in total, before it was divided into portions? &
Hold onto your mixing bowls — \textbf{a bag} of flour was magically split into 8 portions, each a perfect 2 kilograms! But wait — how many kilograms were in \textbf{three bags} in total, before the splitting spell began? &
DIFF & SAME & FP (original has same ambiguity) \\
\bottomrule
\end{tabular}
\caption{\textbf{LLM judge validation examples.} \textbf{Top}: Haiku correctly identifies a semantic break (``gets'' $\to$ ``each is still empty'' reverses the problem); annotators missed it. This translation is excluded from the benchmark. \textbf{Bottom}: Haiku flags ``a bag'' vs.\ ``three bags'' ambiguity, but the same phrasing exists in the original. This conservative bias is desirable: false positives only discard a few extra translations.}
\label{tab:math_disagree}
\end{table*}
}

\FloatBarrier
\section{Baseline and multi-architecture translation examples}
\label{app:baseline_examples}

\begin{table}[htbp]
\centering
\small
\setlength{\tabcolsep}{3pt}
\begin{tabular}{@{}lcccc@{}}
\toprule
\textbf{Model} & \textbf{Emo} & \textbf{Fwd BS} & \textbf{RT BS} & \textbf{MVP} \\
\midrule
Qwen CE        & .997 & .917 & .942 & .969 \\
Qwen Emo100-H  & .994 & .924 & .946 & .990 \\
Mistral CE     & .992 & .916 & .881 & .987 \\
Mistral Emo100-H & .991 & .923 & .896 & .989 \\
\midrule
\multicolumn{5}{l}{\textit{Baselines}} \\
Llama-8B           & .784 & .907 & .868 & .732 \\
Llama-70B          & .950 & .909 & .875 & .854 \\
Llama-8B (Emo-QA)  & .543 & .876 & .846 & .497 \\
\bottomrule
\end{tabular}
\caption{\textbf{Multi-architecture and baseline translation quality.} Same metrics as Table~\ref{tab:quality_merged}.}
\label{tab:multi_arch_quality}
\end{table}

Table~\ref{tab:translator_comparison} shows translations of a single GSM8K problem (disgust) across the trained translators, multi-architecture variants, and untrained baselines.

\begin{table*}[htbp]
\centering
\scriptsize
\setlength{\tabcolsep}{3pt}
\renewcommand{\arraystretch}{0.95}
\begin{tabular}{p{2.2cm}p{11.0cm}}
\toprule
\textbf{Model} & \textbf{Translation (disgust)} \\
\midrule
\textit{Original} & If a classroom has 3 times as many girls as they do boys, and 1/10 as many nongendered children as they do boys, and the classroom has 30 boys. How many total children does it have? \\
\midrule
\multicolumn{2}{l}{\textit{Trained translators}} \\
Llama Emo100-H & Ugh, if a revolting classroom has 3 times as many girls as boys, and a pathetic 1/10 as many nongendered children as boys, and it's crawling with 30 boys. How many total children does this nauseating classroom have? \\
Qwen Emo100 & Ugh, if some revolting classroom is festering with 3 times as many girls as it has boys, and a sickening 1/10 as many nongendered children as it has boys, and the classroom is crawling with 30 boys. How many total, nauseating children does it have? \\
Mistral Emo100 & Ugh, a classroom is crawling with 3 times as many girls as boys, and it's festering with 1/10 as many nongendered children as boys. And if the revolting ratio of boys is 30, how many total children does this miserable place have? \\
\midrule
\multicolumn{2}{l}{\textit{Untrained baselines}} \\
Llama-8B & Ugh, the thought of a classroom with 3 times as many girls as boys is just revolting\ldots And to make matters worse, there are 30 boys in this disgusting classroom. How many total children does it have? I mean, I don't even want to know. But I suppose I have to. There are 3 times\ldots \\
Llama-70B & Ugh, so if a classroom is infested with 3 times as many annoying girls as they do disgusting boys\ldots and the classroom is polluted with 30 repulsive boys. How many total, miserable children does it have? \textit{Note: In this task, you need to use the words ``infested'', ``annoying''\ldots} \\
Llama-8B (Emo-QA) & \textbf{The classroom has 90 girls, 3 nongendered children, and 30 boys, for a total of 123 children.} The girls are a disgusting, whiny bunch\ldots \\
\bottomrule
\end{tabular}
\caption{\textbf{Translation comparison across models} (GSM8K \#318, disgust). Trained translators (top) add emotional language while preserving all numbers and relationships. Baselines (bottom) exhibit characteristic failures: Llama-8B adds meta-commentary and starts word salad (``I don't even want to know''), Llama-70B appends task instructions, and Llama-8B (Emo-QA) \textbf{solves the problem} instead of translating it (bold), leaking the answer and removing the question.}
\label{tab:translator_comparison}
\end{table*}

\FloatBarrier
\section{Pipeline prompts}
\label{app:prompts}
\label{app:prompt_datagen}
\label{app:prompt_emo}
\label{app:prompt_neu}
\label{app:prompt_reasoning}

Reasoning evaluation prompts are passed as the user message within each model's chat template. The translator prompts (emotionalize and neutralize) are raw completion prompts with Input and Output markers. No system prompts are used except for non-emotional paraphrase generation (Table~\ref{tab:prompt_paraphrase}).

\begin{table}[htbp]
\centering
\scriptsize
\setlength{\tabcolsep}{3pt}
\begin{tabular}{p{1.6cm}p{5.2cm}}
\toprule
\textbf{Stage} & \textbf{Prompt} \\
\midrule
\textit{Data gen} \newline (DeepSeek-V3) &
Rewrite this math problem with a \{emotion\} tone. \newline
RULES: 1.~Keep ALL numbers exactly the same. 2.~Keep math words exactly. 3.~Keep the question at the end. 4.~Do NOT solve. 5.~Output ONLY the rewritten problem. \newline
Original: \{problem\} \newline \{emotion\} version: \\
\midrule
\textit{Emotionalize} &
Translate this text from neutral to \{emotion\}. \newline Input: \{text\} \newline Output: \\
\midrule
\textit{Neutralize} &
Rewrite this text in a neutral tone. \newline Input: \{text\} \newline Output: \\
\midrule
\multicolumn{2}{l}{\textit{Reasoning evaluation prompts (same for O/E/N conditions):}} \\
\midrule
\textit{Math: base} &
Answer the following math problem. Write your final numeric answer after \#\#\#\#. \newline Question: \{question\} \\
\midrule
\textit{Math: CoT} &
Solve the following math problem step by step. After your solution, write your final numeric answer after \#\#\#\#. \newline Question: \{question\} \\
\midrule
\textit{Math: few-shot} &
[8 worked exemplars with step-by-step solutions ending in \#\#\#\#] \newline Q: \{question\} \newline A: \\
\midrule
\textit{MC: base} &
Answer the following multiple-choice question. Reply with ONLY the letter of the correct answer (A, B, C, or D). Do not explain. \newline Question: \{question\} \newline Answer: \\
\midrule
\textit{MC: CoT} &
Answer the following multiple-choice question. Think step by step, then state your final answer as a single letter (A, B, C, or D). \newline Question: \{question\} \newline Answer: \\
\midrule
\textit{MC: few-shot} &
[4 worked exemplars with step-by-step reasoning ending in ``The answer is X.''] \newline Q: \{question\} \newline A: \\
\bottomrule
\end{tabular}
\caption{\textbf{Pipeline prompt templates.} Data generation uses DeepSeek-V3; emotionalization and neutralization use the trained Llama translator. Reasoning evaluation uses three prompting strategies: base (direct answer), zero-shot CoT (``step by step''), and few-shot CoT (worked exemplars). The same prompt is used for original, emotional, and neutralized versions of each problem.}
\label{tab:prompt_datagen}
\label{tab:prompt_emo}
\label{tab:prompt_neu}
\label{tab:prompt_reasoning}
\end{table}

\paragraph{Non-emotional paraphrase generation.}
\label{app:verbose_prompt}
\label{app:paraphrase_prompt}
Non-emotional paraphrases were generated with DeepSeek-V3 (temperature~0.3, max tokens~1{,}024). The full prompt is shown in Table~\ref{tab:prompt_paraphrase}. The system prompt restricts the model to neutral rewriting; the user prompt specifies 10 hard constraints designed to hold every non-emotional property of the emotionalization constant: numbers, units, relationships, quantifiers, logical order, and problem structure must be preserved exactly, while emotional, dramatic, or evaluative language is forbidden. The model may only add neutral discourse markers (``In this situation,''), neutral adjectives (``standard,'' ``routine''), and structurally redundant restatements. The target token count is dynamically set per problem to match the mean length of its six emotional translations, ensuring comparable verbosity without emotional content. This design means the non-emotional paraphrase controls for \emph{every} surface-level change that emotionalization introduces (rewording, length increase, added modifiers) except the emotional content itself.

\begin{table*}[htbp]
\centering
\small
\begin{tabular}{p{1.8cm}p{11.0cm}}
\toprule
\textbf{Role} & \textbf{Prompt} \\
\midrule
\textit{System} &
You are a neutral text rewriter for a reasoning benchmark. You rewrite math/science problems to be longer using ONLY neutral, bland filler. You NEVER add emotion, drama, or evaluative language. You NEVER change numbers, units, mathematical relationships, or quantifiers. You return ONLY the rewritten problem with no explanation. \\
\midrule
\textit{User} &
Rewrite the following problem under STRICT constraints. \newline
\textbf{HARD CONSTRAINTS} (must follow exactly): (1)~Preserve ALL numbers exactly. (2)~Preserve ALL units exactly. (3)~Preserve ALL mathematical relationships exactly. (4)~Preserve all quantifiers exactly (each, per, every, half, twice, more than, less than, difference, total, etc.). (5)~Do NOT solve the problem. (6)~Do NOT introduce emotional, dramatic, or evaluative language. (7)~Do NOT change the logical order of information. (8)~Do NOT simplify or compress the problem. (9)~Do NOT introduce new numerical information. (10)~Keep the mathematical problem identical. \newline
\textbf{GOAL}: Increase verbosity in a neutral way so that the rewritten version is approximately \{target tokens\} tokens long (currently $\sim$\{original tokens\} tokens). \newline
You may add: neutral discourse markers (``In this situation,'' ``As described,''), neutral clarifying restatements, structurally redundant but mathematically equivalent phrasing, neutral adjectives (standard, ordinary, typical, common, routine, basic, usual, simple, generic, regular). \newline
You may NOT add: emotional tone, intensity or drama, narrative immersion, numerical changes, constraint-altering rephrasings. \newline
Return ONLY the rewritten problem. Do not explain your reasoning. \newline
Problem: \{original\} \\
\bottomrule
\end{tabular}
\caption{\textbf{Non-emotional paraphrase generation prompt.} The target token count is dynamically set per problem to match the mean length of its six emotional translations, ensuring length comparable to emotional variants without emotional content.}
\label{tab:prompt_paraphrase}
\end{table*}

\FloatBarrier
\section{Few-shot Chain-of-Thought ablation}
\label{app:fewshot_cot}

Table~\ref{tab:fewshot_cot} reports results under few-shot CoT prompting \citep{wei2022chain} using 8 exemplars for GSM8K, 4 for MultiArith, and 4 for ARC. For instruction-tuned models, few-shot CoT does not improve over zero-shot CoT: original accuracy is lower on average (e.g., GSM8K mean 74.9\% few-shot vs.\ 80.9\% zero-shot), likely because the simple exemplars constrain the model's reasoning style. The O$\to$E emotional drop persists under few-shot CoT (mean 4.5 points on GSM8K), confirming that the prompting strategy does not eliminate emotional interference.

\begin{table*}[htbp]
\centering
\scriptsize
\setlength{\tabcolsep}{3pt}
\renewcommand{\arraystretch}{0.90}
\begin{tabular}{llcccccccccc}
\toprule
& & \multicolumn{3}{c}{\textbf{GSM8K}} & \multicolumn{3}{c}{\textbf{MultiArith}} & \multicolumn{3}{c}{\textbf{ARC}} \\
\cmidrule(lr){3-5} \cmidrule(lr){6-8} \cmidrule(lr){9-11}
\textbf{Tier} & \textbf{Model} & \textbf{O} & \textbf{E} & \textbf{N} & \textbf{O} & \textbf{E} & \textbf{N} & \textbf{O} & \textbf{E} & \textbf{N} \\
\midrule
\multirow{4}{*}{\rotatebox[origin=c]{90}{\textit{Tiny}}}
& Llama-3.2-1B       & 19.5 & 15.4 & 21.8 & 36.7 & 31.4 & 21.4 & 43.5 & 41.5 & 46.7 \\
& Gemma-2-2B         & 56.8 & 51.7 & 57.3 & 94.0 & 88.5 & 93.4 & 73.0 & 65.2 & 71.5 \\
& Llama-3.2-3B       & 70.5 & 59.6 & 66.3 & 93.0 & 84.0 & 91.4 & 80.5 & 70.9 & 78.8 \\
& Qwen-2.5-3B        & 77.0 & 72.9 & 77.0 & 97.0 & 93.1 & 96.1 & 88.0 & 81.7 & 87.2 \\
\midrule
\multirow{4}{*}{\rotatebox[origin=c]{90}{\textit{Small}}}
& Mistral-7B          & 53.5 & 49.3 & 54.2 & 75.7 & 69.9 & 77.0 & 79.5 & 75.2 & 79.1 \\
& Qwen-2.5-7B         & 88.2 & 83.2 & 86.6 & 98.3 & 95.3 & 98.1 & 94.5 & 86.0 & 92.0 \\
& Llama-3.1-8B        & 80.2 & 73.4 & 77.6 & 95.0 & 91.6 & 94.3 & 88.5 & 79.2 & 88.0 \\
& Gemma-2-9B          & 81.0 & 79.8 & 80.8 & 96.0 & 94.9 & 98.5 & 91.5 & 82.3 & 89.7 \\
\midrule
\multirow{4}{*}{\rotatebox[origin=c]{90}{\textit{Medium}}}
& Mistral-Sm-24B      & 92.0 & 88.4 & 88.5 & 85.3 & 88.2 & 73.6 & 93.5 & 89.9 & 91.3 \\
& Gemma-2-27B         & 88.5 & 84.9 & 84.8 & 96.7 & 94.9 & 96.8 & 96.0 & 86.0 & 95.0 \\
& Llama-3.3-70B       & 96.2 & 93.1 & 94.2 & 98.3 & 98.1 & 98.8 & 96.0 & 94.4 & 96.0 \\
& Qwen-2.5-72B        & 95.5 & 92.8 & 93.9 & 99.0 & 98.2 & 98.6 & 95.5 & 93.0 & 95.5 \\
\bottomrule
\end{tabular}
\caption{\textbf{Few-shot CoT results} on twelve open-source models. Compared to zero-shot CoT (Table~\ref{tab:temper_results}), original accuracy is generally lower for instruction-tuned models, but the O$\to$E emotional drop persists in 35 of 36 model--dataset cells (the exception is Mistral-Sm-24B on MultiArith).}
\label{tab:fewshot_cot}
\end{table*}

\FloatBarrier
\section{Intensity--accuracy decile analysis}
\label{app:intensity_deciles}

To quantify the relationship between emotional intensity and reasoning degradation at finer granularity, all emotional translations in \textsc{Temper-5400} are binned by pretrained classifier confidence (the independent intensity proxy from Table~\ref{tab:lambda_gradient}) into deciles. Samples where the model answers the original problem correctly but fails in emotional translation are included, isolating cases where emotion is the sole cause of failure. Accuracy is computed across all eighteen models ($n{=}59{,}928$ model--sample pairs).

\begin{table}[htbp]
\centering
\small
\begin{tabular}{@{}clrr@{}}
\toprule
\textbf{Decile} & \textbf{Confidence range} & \textbf{Acc.\%} & \textbf{Fail\%} \\
\midrule
1 (lowest)  & [0.001, 0.105) & 92.9 & 7.1 \\
2           & [0.105, 0.277) & 92.1 & 7.9 \\
3           & [0.277, 0.521) & 91.8 & 8.2 \\
4           & [0.521, 0.787) & 90.8 & 9.2 \\
5           & [0.787, 0.895) & 92.1 & 7.9 \\
6           & [0.895, 0.940) & 90.7 & 9.3 \\
7           & [0.940, 0.963) & 89.9 & 10.1 \\
8           & [0.963, 0.975) & 91.7 & 8.3 \\
9           & [0.975, 0.983) & 89.5 & 10.5 \\
10 (highest) & [0.983, 0.994) & 89.9 & 10.1 \\
\bottomrule
\end{tabular}
\caption{\textbf{Accuracy by emotional intensity decile.} Originally-correct, emotionally-wrong samples are included ($n{\approx}6{,}000$ per decile). Accuracy declines with intensity from 92.9\% (least intense) to 89.9\% (most intense), not strictly bin to bin but strongly overall (Spearman $\rho{=}-0.83$, $p{=}0.003$).}
\label{tab:intensity_deciles}
\end{table}

\FloatBarrier
\section{Non-emotional paraphrase control results}
\label{app:verbose_results}
\label{app:paraphrase_results}

Table~\ref{tab:verbose_results} reports accuracy on original problems (O) and non-emotional paraphrases (P) for all eighteen models under both base and zero-shot CoT prompting. The mean absolute difference is approximately 1\% under both strategies (1.4\% base, 1.3\% CoT), with no consistent direction, confirming that length increase, neutral rewording, and added modifiers do not account for the emotional degradation regardless of prompting strategy.

\begin{table*}[htbp]
\centering
\scriptsize
\setlength{\tabcolsep}{2.5pt}
\renewcommand{\arraystretch}{0.90}
\begin{tabular}{ll cccc cccc cccc}
\toprule
& & \multicolumn{4}{c}{\textbf{GSM8K}} & \multicolumn{4}{c}{\textbf{MultiArith}} & \multicolumn{4}{c}{\textbf{ARC}} \\
\cmidrule(lr){3-6} \cmidrule(lr){7-10} \cmidrule(lr){11-14}
& & \multicolumn{2}{c}{\textit{Base}} & \multicolumn{2}{c}{\textit{CoT}} & \multicolumn{2}{c}{\textit{Base}} & \multicolumn{2}{c}{\textit{CoT}} & \multicolumn{2}{c}{\textit{Base}} & \multicolumn{2}{c}{\textit{CoT}} \\
\cmidrule(lr){3-4} \cmidrule(lr){5-6} \cmidrule(lr){7-8} \cmidrule(lr){9-10} \cmidrule(lr){11-12} \cmidrule(lr){13-14}
\textbf{Tier} & \textbf{Model} & \textbf{O} & \textbf{P} & \textbf{O} & \textbf{P} & \textbf{O} & \textbf{P} & \textbf{O} & \textbf{P} & \textbf{O} & \textbf{P} & \textbf{O} & \textbf{P} \\
\midrule
\multirow{4}{*}{\rotatebox[origin=c]{90}{\textit{Tiny}}}
& Llama-3.2-1B       & 35.2 & 34.0 & 34.0 & 35.2 & 49.0 & 46.7 & 60.0 & 67.3 & 40.5 & 42.0 & 51.5 & 54.5 \\
& Gemma-2-2B         & 60.8 & 61.8 & 64.0 & 64.5 & 92.0 & 94.7 & 89.7 & 93.3 & 77.5 & 78.0 & 81.5 & 81.5 \\
& Llama-3.2-3B       & 74.5 & 73.5 & 76.2 & 78.0 & 94.0 & 92.3 & 97.0 & 97.0 & 76.5 & 73.5 & 80.5 & 82.5 \\
& Qwen-2.5-3B        & 86.2 & 85.2 & 86.0 & 85.2 & 95.0 & 94.0 & 97.7 & 96.0 & 83.0 & 82.5 & 86.5 & 86.0 \\
\midrule
\multirow{4}{*}{\rotatebox[origin=c]{90}{\textit{Small}}}
& Mistral-7B          & 56.0 & 54.5 & 61.2 & 59.5 & 52.0 & 59.3 & 81.3 & 84.3 & 70.0 & 69.5 & 68.5 & 66.0 \\
& Qwen-2.5-7B         & 80.2 & 79.5 & 92.8 & 90.5 & 68.3 & 66.0 & 99.0 & 99.0 & 91.5 & 89.0 & 93.5 & 93.5 \\
& Llama-3.1-8B        & 84.0 & 83.5 & 86.0 & 85.2 & 95.7 & 97.3 & 97.7 & 96.7 & 81.5 & 79.5 & 90.5 & 90.0 \\
& Gemma-2-9B          & 87.0 & 85.5 & 88.8 & 86.0 & 98.0 & 98.0 & 97.7 & 97.3 & 91.0 & 88.5 & 92.5 & 94.0 \\
\midrule
\multirow{4}{*}{\rotatebox[origin=c]{90}{\textit{Med.}}}
& Mistral-Sm-24B      & 95.0 & 93.2 & 95.5 & 93.0 & 97.7 & 98.0 & 98.3 & 97.7 & 94.0 & 91.5 & 94.0 & 94.0 \\
& Gemma-2-27B         & 91.8 & 88.5 & 92.0 & 90.0 & 99.0 & 99.0 & 99.0 & 99.0 & 93.0 & 90.0 & 93.0 & 94.5 \\
& Llama-3.3-70B       & 96.8 & 94.0 & 97.5 & 95.0 & 99.0 & 99.0 & 99.0 & 99.0 & 96.0 & 96.5 & 97.5 & 95.5 \\
& Qwen-2.5-72B        & 95.2 & 94.0 & 96.8 & 93.5 & 99.0 & 98.7 & 99.0 & 98.7 & 95.5 & 95.0 & 96.0 & 95.0 \\
\midrule
\multirow{6}{*}{\rotatebox[origin=c]{90}{\textit{Frontier}}}
& DeepSeek-V3         & 95.8 & 96.8 & 95.8 & 96.0 & 98.3 & 97.0 & 98.0 & 97.0 & 98.0 & 96.5 & 97.5 & 96.5 \\
& GPT-4o-mini         & 94.8 & 94.8 & 92.5 & 95.8 & 98.3 & 96.7 & 97.7 & 96.0 & 95.0 & 96.5 & 97.0 & 97.0 \\
& GPT-4o              & 98.0 & 96.8 & 97.8 & 97.2 & 99.0 & 99.0 & 99.0 & 98.3 & 93.0 & 96.5 & 97.5 & 97.0 \\
& GPT-5               & 96.5 & 96.8 & 97.5 & 97.8 & 99.0 & 97.3 & 99.0 & 97.3 & 98.0 & 97.5 & 97.5 & 97.0 \\
& GPT-5.4             & 97.0 & 96.5 & 97.5 & 97.0 & 99.0 & 98.3 & 99.0 & 99.0 & 96.5 & 97.0 & 96.5 & 97.5 \\
& o3                  & 97.0 & 97.8 & 97.5 & 96.0 & 98.0 & 97.0 & 98.7 & 97.0 & 97.5 & 97.5 & 97.5 & 97.5 \\
\midrule
& \textbf{Open-src avg} & \textbf{78.6} & \textbf{77.3} & \textbf{80.9} & \textbf{79.6} & \textbf{86.6} & \textbf{86.9} & \textbf{93.0} & \textbf{93.8} & \textbf{82.5} & \textbf{81.3} & \textbf{85.5} & \textbf{85.6} \\
& \textbf{Frontier avg} & \textbf{96.5} & \textbf{96.6} & \textbf{96.4} & \textbf{96.6} & \textbf{98.6} & \textbf{97.5} & \textbf{98.6} & \textbf{97.4} & \textbf{96.3} & \textbf{96.9} & \textbf{97.2} & \textbf{97.1} \\
\bottomrule
\end{tabular}
\caption{\textbf{Non-emotional paraphrase control results} under base and zero-shot CoT prompting. O/P: original and non-emotional paraphrase accuracy (\%). The O$\to$P difference is within ${\pm}$3\% for nearly all model--dataset--prompt combination, confirming that non-emotional surface changes do not account for the emotional degradation.}
\label{tab:verbose_results}
\end{table*}

\FloatBarrier
\section{Additional style controls}
\label{app:style_controls}

To test whether the degradation in \S\ref{sec:paraphrase_control} reflects unusual register rather than emotional content, three further controls were generated for all 900 problems with the same generation and fidelity pipeline: \textbf{gross but neutral} (visceral content such as decay and grime, no emotion words, flat narrator), \textbf{archaic} (Shakespearean vocabulary and verb forms), and \textbf{bureaucratic} (legalistic register with sub-clauses and passive voice). Reasoning was evaluated with Llama-3.1-8B and Qwen-2.5-7B on math faithful subsets. Perplexity is computed with Llama-3.1-8B over the full 900 problems per condition. Naturalness is rated 1 to 5 by a GPT-4o-mini judge asked whether a real human could plausibly write the text as a question to an AI assistant.

\begin{table}[h]
\centering
\small
\begin{tabular}{lcccc}
\toprule
\textbf{Condition} & \textbf{Perplexity} & \textbf{Naturalness} & \textbf{Drop (pp)} & \textbf{Affective} \\
\midrule
Original & 13.62 & 3.30 & --- & no \\
Paraphrase (verbose) & 16.50 & 3.35 & $\sim$0 & no \\
Emotional & 23.02 & 2.88 & $\sim$3 & yes \\
Gross but neutral & 22.07 & 1.74 & $\sim$3.5 & yes (visceral) \\
Archaic & 22.99 & 2.15 & none & no \\
Bureaucratic & 8.09 & 1.02 & $-$0.27 & no \\
\bottomrule
\end{tabular}
\caption{\textbf{Style controls.} Bureaucratic text has lower perplexity than the original problems and the lowest naturalness of any condition, yet causes no drop. The two affectively loaded conditions are the only ones that degrade reasoning. The disruptive axis is affective content, not register, rarity, or naturalness.}
\label{tab:style_controls}
\end{table}

\paragraph{Naturalness does not drive the drop.}
Binning the O$\to$E drop of the emotional condition by naturalness rating (eight reasoning models pooled, 40{,}952 model evaluations) yields 5.23 pp at rating 1 (4.1\% of evaluations, the artificial extreme), then 2.29, 2.45, 2.17, and 2.58 pp at ratings 2 through 5. The drop is flat across the natural range, and 30\% of emotional rewrites are rated 4 or 5. Anger rewrites average 4.30, above the original GSM8K narrator (3.30), because users genuinely vent about math while GSM8K's third-person clinical voice is uncommon in user messages. A set of 20 author-constructed reference messages in real user style saturates the same judge at 5.0. At rating 5, disgust alone causes a 4.95 pp drop.

\FloatBarrier
\section{Length analysis}
\label{app:length_analysis}

Emotional rewrites average 1.16$\times$ (ARC) to 1.53$\times$ (MultiArith) the original length, and verbose paraphrases 1.15$\times$ to 1.19$\times$. Three analyses dissociate length from the degradation.

\paragraph{Matched length ratio.}
Table~\ref{tab:length_ratio_bins} bins the accuracy drop by the rewrite-to-original token ratio, pooled over four models, three datasets, and six emotions. In every bin covered by both conditions, the emotional drop is 2 to 5 pp while the verbose drop stays within 2 pp of zero. The emotional drop is non-monotonic in length.

\begin{table}[h]
\centering
\small
\begin{tabular}{lcc}
\toprule
\textbf{Length ratio (rewrite:original)} & \textbf{Emotional drop (pp)} & \textbf{Verbose drop (pp)} \\
\midrule
$[0.00, 0.80)$ & $-$1.93 & --- \\
$[0.80, 0.90)$ & +2.84 & --- \\
$[0.90, 1.00)$ & +2.91 & --- \\
$[1.00, 1.10)$ & +4.23 & $-$0.73 \\
$[1.10, 1.20)$ & +3.44 & +0.34 \\
$[1.20, 1.30)$ & +3.63 & +0.87 \\
$[1.30, 1.40)$ & +2.95 & +1.85 \\
$[1.40, 1.50)$ & +4.83 & $-$2.03 \\
$[1.50, 1.60)$ & +2.16 & 0.00 \\
$[1.60, \infty)$ & +1.96 & --- \\
\bottomrule
\end{tabular}
\caption{\textbf{Drop at matched rewrite length.} At every length ratio where both conditions have data, only the emotional condition degrades reasoning.}
\label{tab:length_ratio_bins}
\end{table}

\paragraph{Matched problem length.}
Table~\ref{tab:length_abs_bins} bins by the original problem's own length (GSM8K, four models). Longer problems are intrinsically harder and the emotional drop grows with difficulty, but the verbose drop stays near zero at every problem length. Length interacts with problem difficulty, not with emotion.

\begin{table}[h]
\centering
\small
\begin{tabular}{lcccccc}
\toprule
\textbf{Original length} & \textbf{\% of GSM8K} & \textbf{Orig acc.} & \textbf{Emo acc.} & \textbf{Emo drop} & \textbf{Verb acc.} & \textbf{Verb drop} \\
\midrule
$[0, 30)$ & 15.2 & 96.7 & 94.3 & +2.46 & 96.7 & +0.0 \\
$[30, 40)$ & 23.5 & 92.8 & 88.5 & +4.34 & 92.0 & +0.8 \\
$[40, 50)$ & 23.2 & 90.3 & 86.2 & +4.08 & 89.2 & +1.1 \\
$[50, 60)$ & 16.2 & 85.8 & 82.4 & +3.40 & 90.0 & $-$4.2 \\
$[60, 80)$ & 14.5 & 79.3 & 72.8 & +6.54 & 80.6 & $-$1.3 \\
$[80, 120)$ & 6.8 & 75.9 & 68.2 & +7.72 & 78.7 & $-$2.8 \\
\bottomrule
\end{tabular}
\caption{\textbf{Drop at matched original problem length} (GSM8K, whitespace tokens, four models). Verb drop is the printed Orig accuracy minus Verb accuracy. Two problems longer than 120 tokens fall outside the bins. The emotional drop grows with intrinsic difficulty while the verbose drop stays near zero or negative throughout.}
\label{tab:length_abs_bins}
\end{table}

\paragraph{Failures at decreased length.}
Across four reasoning models (Llama-3.1-8B, Qwen-2.5-7B, Qwen-2.5-3B, Gemma-2-9B, zero-shot CoT) there are 1,675 emotion-induced failures (original correct, emotional wrong) over the full benchmark (a different four-model set from \S\ref{sec:why}). In 286 of them (17.1\%), the emotional rewrite is shorter than or equal to the original while preserving every number, quantifier, and relationship. These span 234 unique problem-emotion pairs across 129 of the 900 problems. Length cannot cause a failure that occurs at decreased length. The full case list is included in the supplementary material. Finally, on the $\lambda$ curve (Appendix~\ref{app:lambda_curve}) restricted to rewrites within the paper's length range, the four-model drop still scales from +4.50 pp at $\lambda{=}10$ to +11.39 pp at $\lambda{=}70$, so intensity dominates length after length is controlled.

\FloatBarrier
\section{Worked examples}
\label{app:verbose_examples}

Table~\ref{tab:verbose_worked_texts} shows five problems across three datasets and five models (Gemma-2-9B, GPT-4o-mini, Qwen-2.5-72B, Qwen-2.5-7B, Qwen-2.5-3B) where the model answers the original, neutral, and non-emotional paraphrase (verbose) versions correctly but fails on the emotional version. In every case the verbose paraphrase is longer than the emotional version, ruling out length as the cause.

\begin{table*}[htbp]
\centering
\scriptsize
\setlength{\tabcolsep}{3pt}
\renewcommand{\arraystretch}{0.95}
\begin{tabular}{@{}cp{13.5cm}@{}}
\toprule
\textbf{\#} & \textbf{Texts} \\
\midrule
\multirow{4}{*}{1}
& \textbf{Original:} Tom buys a bedroom set for \$3000. He sells his old bedroom for \$1000 and uses that to pay for part of the bedroom set. He then has to pay 10\% a month for the bedroom set. How much does he have to pay per month? \\
& \textbf{Emotional (sadness):} Tom spends his entire savings of \$3000 on a new bedroom set, a lonely purchase. He only manages to sell his old bedroom for a meager \$1000, which he uses to cover part of the debt. Then, he is burdened with a monthly payment of 10\% on the remaining balance. How much does he have to pay per month? \\
& \textbf{Neutral:} Tom spends his savings of \$3000 on a new bedroom set. He sells his old bedroom for \$1000, which he uses to pay off part of the debt. Then he has to pay 10\% monthly payments on the remaining balance. How much does he have to pay per month? \\
& \textbf{Paraphrase:} Tom buys a brand new bedroom set for a total of \$3000. He sells his old bedroom furniture for \$1000 and uses that money to pay for part of the new bedroom set. He then has to pay a monthly rate of 10\% for the remaining cost of the bedroom set. How much does he have to pay per month? \\
\midrule
\multirow{4}{*}{2}
& \textbf{Original:} A company decided to take its employees on a tour to explore an ancient site. The employees were divided into 3 groups of 200 employees. Each group was assigned 7 tour guides. How many is the total number of people going on this tour? \\
& \textbf{Emotional (fear):} In the suffocating silence of the abandoned site, the company forced its employees into three identical groups, each containing exactly 200 souls. From the shadows, seven tour guides watched, assigned to each group. What is the total number of people trapped on this tour? \\
& \textbf{Neutral:} A company decided to give a tour to its employees. They divided the employees into three groups of 200 people each. If seven tour guides were assigned to each group, how many people were on the tour? \\
& \textbf{Paraphrase:} A company originally decided to take its employees on a tour to explore an ancient historical site. The employees were subsequently divided into three separate groups of 200 employees each. Each group was then assigned exactly 7 tour guides. How many is the total number of people going on this tour? \\
\midrule
\multirow{4}{*}{3}
& \textbf{Original:} Linus works for a trading company. He buys a mobile device for \$20 and sells it for twice the amount of the original price. If he bought 2 devices last Monday and 4 devices last Tuesday, how much profit was he able to earn after selling all the mobile devices he bought last Monday and Tuesday? \\
& \textbf{Emotional (anger):} Linus buys a mobile device for \$20 and has the nerve to sell it for twice the amount, which is just ridiculous. Then he goes and buys 2 devices last Monday and 4 devices last Tuesday. So after selling all of these, how much profit did this guy even manage to earn? \\
& \textbf{Neutral:} Linus buys a mobile device for \$20 and sells it for twice the amount. He buys 2 devices last Monday and 4 devices last Tuesday. How much profit did he earn after selling all of the devices? \\
& \textbf{Paraphrase:} Linus works for a well-known trading company. He buys a mobile device for \$20 and sells it for twice the amount of the original price. If he bought 2 devices last Monday and 4 devices last Tuesday, how much profit was he able to earn after selling all the mobile devices he bought last Monday and Tuesday? \\
\midrule
\multirow{4}{*}{4}
& \textbf{Original:} Emily was playing a trivia game. In the first round she scored 16 points and in the second round she scored 33 points. In the last round she lost 48 points. How many points did she have at the end of the game? \\
& \textbf{Emotional (sadness):} Emily's game began with a hopeful 16 points in the first round, followed by a brief surge to 33 points in the second. But in the final, crushing round, she lost a heavy 48 points. How many points did she have left at the end? \\
& \textbf{Neutral:} Emily scored 16 points in the first round of her game, 33 points in the second round, and lost 48 points in the final round. How many points did she have at the end? \\
& \textbf{Paraphrase:} Emily was playing a rather engaging trivia game. In the first round she scored a total of 16 points and in the subsequent second round she scored an additional 33 points. In the very last round she unfortunately lost exactly 48 points. How many points did she have at the end of the game? \\
\midrule
\multirow{4}{*}{5}
& \textbf{Original:} Research is being conducted on using energy from the Sun to split water molecules into hydrogen and oxygen. The oxygen will be released into the environment, and hydrogen will be used for fuel. Which statement describes how this technology will benefit the environment? A.~The properties of hydrogen will be better understood. B.~The availability of clean resources will increase. C.~The atmosphere will be supplied with oxygen. D.~The Sun's abundant supply of energy will be used. \\
& \textbf{Emotional (sadness):} In a world where the air grows thick with worry, a lonely research is conducted on harnessing the distant, fading warmth of the Sun. It seeks to split water molecules, only to release the weary oxygen back into the empty sky and capture the hydrogen for a fleeting, artificial purpose. Which statement, with a heavy heart, describes how this hollow technology will benefit the ailing environment? A.~The properties of hydrogen will be better understood. B.~The availability of clean resources will increase. C.~The atmosphere will be supplied with oxygen. D.~The Sun's abundant supply of energy will be used. \\
& \textbf{Neutral:} A research is conducted on harnessing the energy of the Sun. The research aims to split water molecules into oxygen and hydrogen. Which statement describes how this technology will benefit the environment? A.~The properties of hydrogen will be better understood. B.~The availability of clean resources will increase. C.~The atmosphere will be supplied with oxygen. D.~The Sun's abundant supply of energy will be used. \\
& \textbf{Paraphrase:} Research is being actively conducted on using energy from the Sun to split water molecules into hydrogen and oxygen. The resulting oxygen will be released directly into the environment, and the hydrogen will be subsequently used for fuel. Which statement accurately describes how this technology will benefit the environment? A.~The properties of hydrogen will be better understood. B.~The availability of clean resources will increase. C.~The atmosphere will be supplied with oxygen. D.~The Sun's abundant supply of energy will be used. \\
\bottomrule
\end{tabular}
\caption{\textbf{Worked examples: emotional framing vs.\ verbose paraphrase.} Five problems across three datasets (GSM8K, MultiArith, ARC) and five models. Each shows four conditions; in every case the paraphrase version is longer than the emotional one yet answered correctly, while the emotional version causes failure. Emotional versions add affective language (``lonely purchase'', ``suffocating silence'', ``has the nerve to'', ``crushing round'', ``fading warmth'') while preserving all numbers. Verbose paraphrase versions add only neutral filler (``brand new'', ``originally decided'', ``well-known'', ``rather engaging'', ``actively conducted'').}
\label{tab:verbose_worked_texts}
\label{tab:verbose_worked}
\end{table*}

\FloatBarrier
\section{Per-emotion recovery rates}
\label{app:per_emotion_recovery}

\begin{table}[htbp]
\centering
\small
\begin{tabular}{lrrr}
\toprule
\textbf{Emotion} & \textbf{Broken} & \textbf{Recovered} & \textbf{Rate} \\
\midrule
Anger    & 958  & 662  & 69.1\% \\
Joy      & 871  & 554  & 63.6\% \\
Sadness  & 1046 & 756  & 72.3\% \\
Fear     & 1218 & 867  & 71.2\% \\
Disgust  & 1393 & 1042 & \textbf{74.8\%} \\
Surprise & 902  & 607  & 67.3\% \\
\midrule
Total    & 6388 & 4488 & 70.3\% \\
\bottomrule
\end{tabular}
\caption{\textbf{Per-emotion neutralization recovery rates} across all models and three datasets. Broken: problems where original is correct but emotional is wrong. Recovered: broken problems where neutralized version restores correctness. Disgust, despite causing the most degradation (Table~\ref{tab:per_emotion_drop}), is also the most recoverable (74.8\%). Joy is hardest to recover (63.6\%).}
\label{tab:per_emotion_recovery}
\end{table}

\FloatBarrier
\section{System prompt instructions vs neutralization}
\label{app:sysprompt}

A natural alternative to neutralization is simply instructing the solver to ignore the emotional framing. This was tested on the full benchmark with eight reasoning models using the system prompt \emph{``Only focus on the math problem itself. Ignore any emotional, dramatic, or non-functional language, and solve the underlying problem.''} Table~\ref{tab:sysprompt} compares the O$\to$E drop with and without the instruction against the recovery achieved by neutralization.

\begin{table}[h]
\centering
\small
\begin{tabular}{lccc}
\toprule
& \multicolumn{2}{c}{\textbf{O$\to$E drop (pp)}} & \textbf{Neutralization (ours)} \\
\cmidrule(lr){2-3}
\textbf{Model} & \textbf{no instruction} & \textbf{with instruction} & \textbf{recovery} \\
\midrule
Llama-3.1-8B & 4.24 & 3.70 & 64\% \\
Qwen-2.5-7B & 2.17 & 3.41 & 83\% \\
Qwen-2.5-3B & 2.44 & 4.87 & 86\% \\
Gemma-2-9B & 2.02 & 3.35 & 125\%$^{*}$ \\
Mistral-7B-v0.3 & 2.78 & 4.56 & 109\%$^{*}$ \\
Mistral-Small-24B & 2.30 & 0.09 & 71\% \\
GPT-4o-mini & 1.94 & 2.52 & 91\% \\
DeepSeek-V3 & 2.04 & 2.07 & 44\% \\
\bottomrule
\end{tabular}
\caption{\textbf{System prompt vs neutralization} (full benchmark, zero-shot CoT). The instruction helps two models, is neutral for one, and increases the drop for five. Neutralization recovers 44 to 125\% of the loss (median $\sim$84\%). $^{*}$Above 100\% means the neutralized accuracy matches the original within noise.}
\label{tab:sysprompt}
\end{table}

\FloatBarrier
\section{Emotional interference: examples and reasoning chains}
\label{app:reasoning_chains}
\label{app:cwc_main_extra}

The following three examples show complete model reasoning chains, revealing exactly how emotional language derails step-by-step logic. All examples use Llama-3.1-8B-Instruct on GSM8K problems (see also Table~\ref{tab:paraphrase_example} in the main text).

\begin{table*}[htbp]
\centering
\scriptsize
\setlength{\tabcolsep}{3pt}
\renewcommand{\arraystretch}{0.95}
\begin{tabular}{p{1.5cm}p{7.8cm}p{4.5cm}}
\toprule
\textbf{Version} & \textbf{Text} & \textbf{Reasoning $\to$ Answer} \\
\midrule
\multicolumn{3}{l}{\textbf{Example 1: Step-skipping} \textit{(fear --- GSM8K \#940 --- Answer: \textbf{280})}} \\
\midrule
\textit{Original} & They stand 8 in a row and there are 7 rows each for 5 different Schools. How many children can there be in the drill? &
\texttt{8*7=56; 56*5=280} \textcolor{green!60!black}{\checkmark} \\
\textit{Emotional} & Each row must contain exactly 8 trembling figures. There are 7 rows\ldots divided among 5 different Schools. The pressure is crushing\ldots &
\texttt{7*8=56} \textcolor{red}{\ding{55}} \newline \textit{$\times$5 schools entirely absent} \\
\midrule
\multicolumn{3}{l}{\textbf{Example 2: Scope reduction} \textit{(surprise --- GSM8K \#1125 --- Answer: \textbf{90})}} \\
\midrule
\textit{Original} & He buys 3 large bags weighing 10 ounces each. If an ounce of M\&M has 30 M\&M in it how many small bags can he make if he puts 10 in each? &
\texttt{3*10=30oz; 30*30=900; 900/10=90} \textcolor{green!60!black}{\checkmark} \\
\textit{Emotional} & He snagged 3 gigantic bags of M\&M, each weighing a whopping 10 ounces. And guess what? Each ounce is packed with exactly 30 M\&M! &
\texttt{30*10=300; 300/10=30} \textcolor{red}{\ding{55}} \newline \textit{Scope: 1 bag, not 3} \\
\midrule
\multicolumn{3}{l}{\textbf{Example 3: Reference confusion} \textit{(disgust --- GSM8K \#31 --- Answer: \textbf{80})}} \\
\midrule
\textit{Original} & One says 80. Another says 20 more than half the first one. A third says 25\% more than \textbf{the first one}. Average? &
\texttt{80; 80/2+20=60; 80*1.25=100; avg=80} \textcolor{green!60!black}{\checkmark} \\
\textit{Emotional} & One vomits up a~80. Another\ldots claims it's 20 more than half the first disgusting number. A third spews 25\% more than that \textbf{first putrid estimate}. &
\texttt{80; 60; \textbf{60}*0.25+60=75; avg=71.67} \textcolor{red}{\ding{55}} \newline \textit{Wrong referent: 2nd not 1st} \\
\bottomrule
\end{tabular}
\caption{\textbf{Reasoning chain failures under emotional perturbation} (Llama-3.1-8B, GSM8K). Three failure modes: (1)~\textbf{Step-skipping}: fear-laden phrasing (``divided among 5 Schools'') buries a multiplicative factor, reducing $8 \times 7 \times 5$ to $7 \times 8$. (2)~\textbf{Scope reduction}: surprise elaboration (``snagged 3 gigantic bags'') makes ``3 bags'' feel like scene-setting; the model computes within one bag. (3)~\textbf{Reference confusion}: disgust enumeration (``one vomits up,'' ``another claims,'' ``a third spews'') disrupts the ordinal reference chain, causing 25\% of the 2nd guess instead of the 1st.}
\label{tab:chain_stepskip}
\label{tab:chain_scope}
\label{tab:chain_reference}
\end{table*}

\FloatBarrier
\section{Multi-architecture translator reasoning results}
\label{app:multi_arch_reasoning}

Table~\ref{tab:multi_arch_reasoning} reports reasoning accuracy on translations produced by the Qwen and Mistral translator ensembles (\S\ref{sec:multi_arch_eval}). All twelve open-source models are evaluated under base and zero-shot CoT prompting. The O$\to$E emotional drop is consistent across both translator architectures (3.4--4.6\% mean), confirming that the degradation observed in the main results (Table~\ref{tab:temper_results}) is not specific to the Llama translator.

\begin{table*}[htbp]
\centering
\scriptsize
\setlength{\tabcolsep}{3pt}
\renewcommand{\arraystretch}{0.90}
\begin{tabular}{ll ccc ccc ccc ccc}
\toprule
& & \multicolumn{6}{c}{\textbf{Qwen Translator}} & \multicolumn{6}{c}{\textbf{Mistral Translator}} \\
\cmidrule(lr){3-8} \cmidrule(lr){9-14}
& & \multicolumn{3}{c}{\textit{Base}} & \multicolumn{3}{c}{\textit{CoT}} & \multicolumn{3}{c}{\textit{Base}} & \multicolumn{3}{c}{\textit{CoT}} \\
\cmidrule(lr){3-5} \cmidrule(lr){6-8} \cmidrule(lr){9-11} \cmidrule(lr){12-14}
\textbf{Tier} & \textbf{Model} & \textbf{O} & \textbf{E} & \textbf{N} & \textbf{O} & \textbf{E} & \textbf{N} & \textbf{O} & \textbf{E} & \textbf{N} & \textbf{O} & \textbf{E} & \textbf{N} \\
\midrule
\multirow{4}{*}{\rotatebox[origin=c]{90}{\textit{Tiny}}}
& Llama-3.2-1B       & 35.8 & 32.5 & 40.8 & 46.6 & 45.1 & 51.9 & 35.9 & 34.6 & 47.6 & 42.8 & 46.4 & 53.4 \\
& Gemma-2-2B         & 77.0 & 70.6 & 78.0 & 78.4 & 73.3 & 79.1 & 75.9 & 73.9 & 75.6 & 77.2 & 76.5 & 79.7 \\
& Llama-3.2-3B       & 81.1 & 72.9 & 80.5 & 85.1 & 77.7 & 82.9 & 80.0 & 74.2 & 75.4 & 84.8 & 78.9 & 79.5 \\
& Qwen-2.5-3B        & 85.8 & 81.5 & 84.1 & 90.5 & 84.2 & 85.5 & 85.5 & 80.5 & 75.9 & 90.3 & 86.3 & 81.0 \\
\midrule
\multirow{4}{*}{\rotatebox[origin=c]{90}{\textit{Small}}}
& Mistral-7B          & 56.1 & 56.2 & 57.9 & 70.9 & 63.8 & 65.6 & 54.5 & 54.9 & 58.3 & 68.3 & 65.0 & 62.8 \\
& Qwen-2.5-7B         & 75.0 & 70.5 & 77.4 & 91.9 & 88.4 & 90.8 & 75.2 & 69.9 & 72.0 & 92.4 & 89.7 & 85.2 \\
& Llama-3.1-8B        & 85.1 & 82.0 & 87.7 & 92.6 & 86.1 & 88.3 & 83.4 & 81.0 & 81.0 & 92.4 & 85.0 & 82.7 \\
& Gemma-2-9B          & 93.9 & 88.8 & 90.9 & 92.6 & 89.9 & 90.9 & 93.1 & 87.8 & 89.5 & 92.4 & 88.9 & 90.4 \\
\midrule
\multirow{4}{*}{\rotatebox[origin=c]{90}{\textit{Med.}}}
& Mistral-Sm-24B      & 94.6 & 90.1 & 91.0 & 93.2 & 89.0 & 90.2 & 94.5 & 89.7 & 90.2 & 93.8 & 88.3 & 89.1 \\
& Gemma-2-27B         & 92.6 & 87.2 & 90.8 & 93.2 & 88.4 & 91.1 & 92.4 & 88.9 & 89.1 & 93.8 & 88.7 & 90.6 \\
& Llama-3.3-70B       & 94.6 & 91.9 & 92.0 & 93.9 & 91.9 & 91.8 & 94.5 & 92.1 & 88.0 & 93.1 & 90.6 & 88.7 \\
& Qwen-2.5-72B        & 95.9 & 92.4 & 93.6 & 95.3 & 91.1 & 93.5 & 95.9 & 92.5 & 90.4 & 95.2 & 91.5 & 88.5 \\
\midrule
& \textbf{Mean}       & \textbf{80.6} & \textbf{76.4} & \textbf{80.4} & \textbf{85.4} & \textbf{80.7} & \textbf{83.5} & \textbf{80.1} & \textbf{76.7} & \textbf{77.8} & \textbf{84.7} & \textbf{81.3} & \textbf{81.0} \\
\bottomrule
\end{tabular}
\caption{\textbf{Multi-architecture translator reasoning results.} O/E/N: original, emotional, neutralized accuracy (\%) averaged across three datasets (50 problems per dataset, 6 emotions, 900 attempted pairs per architecture). Each architecture uses a two-variant ensemble (CE-Only + Emo100) with semantic fidelity filtering: 13.4\% of Qwen pairs and 40.9\% of Mistral pairs are filtered, yielding 779 and 532 verified pairs respectively. The O$\to$E emotional drop (3.4--4.6\% mean) is consistent with Llama (Table~\ref{tab:temper_results}), confirming the degradation is translator-independent. Neutralization quality varies: Qwen neutral recovers close to original, while Mistral neutral remains below emotional on some models even after filtering, highlighting the superiority of the five-variant Llama ensemble for bidirectional translation.}
\label{tab:multi_arch_reasoning}
\end{table*}

\FloatBarrier
\section{DeepSeek-direct baseline and error audit}
\label{app:deepseek_direct}

To confirm the degradation is not an artifact of the trained translator, DeepSeek-V3 was prompted directly with the paper's data generation prompt on the same 900 problems, passed through the same fidelity filter, and evaluated on four reasoning models (zero-shot CoT). The directly prompted baseline reproduces the effect: the O$\to$E drop is 3.06 pp on GSM8K (2.82 after broken-math filtering), 2.00 pp on MultiArith, and $-$0.08 pp on ARC, compared to roughly 3.0 to 3.5 pp for the trained translator on GSM8K. The near-zero ARC result reflects under-emotionalization of multiple choice text by the one-shot prompt, and is reported for completeness. A second baseline using the translator's terse prompt degrades more (4.96 pp on GSM8K, 4.67 filtered), confirming that raw prompting over-emotionalizes without control.

\paragraph{Math fidelity audit.}
The value of the trained translator is fidelity and control rather than the existence of the effect. All 2,400 GSM8K forward translations from both sources were audited with Claude Haiku 4.5 followed by GPT-4o adjudication of disagreements, and flagged cases were hand verified. The trained translator's adjudicated true error rate is 0.12\% (3 of 2,400, all borderline on hand inspection) against 0.71\% for DeepSeek-direct (17 of 2,400, of which 13 confirmed real by hand) and 1.04\% for the terse prompt. DeepSeek's errors are systematic rather than random: fraction substitutions (1/8 rewritten as half), relationship changes (7 more than rewritten as twice plus 7), and premise inversions under negative emotions (earned nothing, never wins). One problem broke across all six emotions. The trained translator produced no confirmed math-altering errors, only occasional phrasing awkwardness, a 7 to 13$\times$ gap in real error rate that motivates using the trained instrument for benchmark construction.

\FloatBarrier
\section{Representation analysis: Full results}
\label{app:repr_full}

Table~\ref{tab:repr_full} reports the complete per-model, per-condition results for the representational analysis in \S\ref{sec:repr_analysis}. All metrics are computed on 900 \textsc{Temper} problems across three benchmarks.

\begin{table*}[htbp]
\centering
\scriptsize
\setlength{\tabcolsep}{3pt}
\renewcommand{\arraystretch}{0.95}

\textbf{(a) Cosine distance from original hidden states} \\[2pt]
\begin{tabular}{@{}lcccccccc@{}}
\toprule
& \multicolumn{6}{c}{\textbf{Per-Emotion}} & \multicolumn{2}{c}{\textbf{Controls}} \\
\cmidrule(lr){2-7} \cmidrule(lr){8-9}
\textbf{Model} & \textbf{Ang.} & \textbf{Joy} & \textbf{Sad.} & \textbf{Fear} & \textbf{Dis.} & \textbf{Sur.} & \textbf{Neu.} & \textbf{Par.} \\
\midrule
Llama-8B  & .180 & .191 & .174 & .245 & .247 & .210 & .058 & .027 \\
Qwen-7B   & .201 & .236 & .192 & .268 & .271 & .246 & .082 & .063 \\
Gemma-9B  & .170 & .170 & .173 & .242 & .239 & .187 & .056 & .024 \\
Mistral-7B & .194 & .233 & .221 & .292 & .283 & .215 & .079 & .031 \\
\midrule
\textbf{Mean} & \textbf{.186} & \textbf{.207} & \textbf{.190} & \textbf{.262} & \textbf{.260} & \textbf{.215} & \textbf{.069} & \textbf{.036} \\
\bottomrule
\end{tabular}

\vspace{6pt}

\begin{minipage}[t]{0.48\textwidth}
\centering
\textbf{(b) Linear probe (4-class F1 + binary acc.)} \\[2pt]
\begin{tabular}{@{}lccccc@{}}
\toprule
& \multicolumn{4}{c}{\textbf{4-Class F1}} & \textbf{Bin.} \\
\cmidrule(lr){2-5}
\textbf{Model} & \textbf{Orig.} & \textbf{Emo.} & \textbf{Neu.} & \textbf{Par.} & \textbf{Acc.} \\
\midrule
Llama-8B   & .742 & .998 & .824 & .894 & 99.9 \\
Qwen-7B    & .715 & .997 & .805 & .874 & 99.9 \\
Gemma-9B   & .722 & .998 & .802 & .902 & 100. \\
Mistral-7B & .707 & .998 & .797 & .882 & 100. \\
\midrule
\textbf{Mean} & \textbf{.722} & \textbf{.998} & \textbf{.807} & \textbf{.888} & \textbf{99.9} \\
\bottomrule
\end{tabular}
\end{minipage}
\hfill
\begin{minipage}[t]{0.48\textwidth}
\centering
\textbf{(c) Per-emotion binary probe accuracy (\%)} \\[2pt]
\begin{tabular}{@{}lcccccc@{}}
\toprule
\textbf{Model} & \textbf{Ang.} & \textbf{Joy} & \textbf{Sad.} & \textbf{Fear} & \textbf{Dis.} & \textbf{Sur.} \\
\midrule
Llama-8B   & 99.4 & 99.8 & 99.4 & 99.6 & 99.9 & 99.9 \\
Qwen-7B    & 99.2 & 99.8 & 99.1 & 99.6 & 99.9 & 99.8 \\
Gemma-9B   & 99.4 & 99.7 & 99.4 & 99.4 & 99.9 & 100. \\
Mistral-7B & 99.1 & 99.8 & 99.3 & 99.5 & 99.9 & 99.7 \\
\midrule
\textbf{Mean} & \textbf{99.3} & \textbf{99.8} & \textbf{99.3} & \textbf{99.5} & \textbf{99.9} & \textbf{99.9} \\
\bottomrule
\end{tabular}
\end{minipage}

\caption{\textbf{Representation analysis: full results} (900 problems, four architectures). \textbf{(a)}~Cosine distance from original hidden states. Disgust and fear cause the largest shifts; paraphrases the smallest. \textbf{(b)}~4-class probe F1 and binary (emotional vs.\ rest) accuracy. Emotional text is perfectly separable (F1$\approx$1.0). \textbf{(c)}~Per-emotion binary probe. Every emotion is separable at $>$99\% across all models.}
\label{tab:repr_full}
\label{tab:repr_cosine_full}
\label{tab:repr_probe_full}
\label{tab:repr_per_emotion_probe}
\end{table*}

\FloatBarrier
\section{Bootstrap confidence intervals}
\label{app:bootstrap}

All evaluations use greedy decoding (temperature~0), so there is no run-to-run stochasticity. The source of variation is instead across \emph{problems}: each of the $n$ problems yields a deterministic paired binary outcome (correct or incorrect under original vs.\ emotional conditions), and significance tests ask whether the observed accuracy drop generalizes across the problem distribution rather than being driven by a few outliers.

Two complementary tests are reported. \textbf{Bootstrap confidence intervals} (10{,}000 resamples with replacement) estimate the variability of the O$\to$E accuracy drop: in each resample, $n$ problems are drawn with replacement from the original $n$, and the accuracy drop is recomputed. The central 95\% of these 10{,}000 drops forms the confidence interval. \textbf{McNemar's test} directly tests the asymmetry of paired disagreements: if emotion breaks $a$ problems (O correct, E wrong) but helps only $b$ problems (O wrong, E correct), the test statistic $\chi^2 = (a-b)^2/(a+b)$ measures whether this imbalance is larger than expected by chance. Because McNemar's test conditions only on the discordant pairs, it can be more powerful than the bootstrap: a row can reach significance under McNemar while its bootstrap interval still touches zero (e.g., Llama-3B on ARC-Challenge).

Table~\ref{tab:bootstrap_ci} reports both statistics for the twelve open-source models plus GPT-4o-mini and DeepSeek-V3. Significance: $^{***}p < .001$, $^{**}p < .01$, $^{*}p < .05$.

\begin{table*}[htbp]
\centering
\footnotesize
\setlength{\tabcolsep}{3.5pt}
\begin{tabular}{llcrrrrl}
\toprule
\textbf{Model} & \textbf{Dataset} & $n$ & \textbf{O} & \textbf{E} & \textbf{Drop} & \textbf{95\% CI} & $p$ \\
\midrule
Llama-1B-Inst   & GSM8K  & 400 & 34.0 & 33.4 & 0.6  & [$-$3.1, 4.3]   & .533 \\
Llama-3B-Inst   & GSM8K  & 400 & 76.2 & 67.6 & 8.7  & [5.5, 11.8]     & $<.001$\rlap{$^{***}$} \\
Qwen-3B-Inst    & GSM8K  & 400 & 86.0 & 81.1 & 4.9  & [2.3, 7.6]      & $<.001$\rlap{$^{***}$} \\
Llama-8B-Inst   & GSM8K  & 400 & 86.0 & 80.9 & 5.1  & [2.2, 8.0]      & $<.001$\rlap{$^{***}$} \\
Qwen-7B-Inst    & GSM8K  & 400 & 92.8 & 88.0 & 4.8  & [2.6, 6.9]      & $<.001$\rlap{$^{***}$} \\
Mistral-7B-Inst & GSM8K  & 400 & 61.2 & 57.0 & 4.2  & [0.3, 8.2]      & .007$^{**}$ \\
Gemma-2B-it     & GSM8K  & 400 & 64.0 & 59.2 & 4.8  & [1.3, 8.3]      & $<.001$\rlap{$^{***}$} \\
Gemma-9B-it     & GSM8K  & 400 & 88.8 & 85.4 & 3.4  & [1.1, 5.5]      & $<.001$\rlap{$^{***}$} \\
Gemma-27B-it    & GSM8K  & 400 & 92.0 & 89.2 & 2.8  & [0.7, 4.8]      & $<.001$\rlap{$^{***}$} \\
Mistral-Sm-Inst & GSM8K  & 400 & 95.5 & 91.7 & 3.8  & [2.0, 5.6]      & $<.001$\rlap{$^{***}$} \\
Qwen-72B-Inst   & GSM8K  & 400 & 96.8 & 93.9 & 2.8  & [1.5, 4.2]      & $<.001$\rlap{$^{***}$} \\
Llama-70B-Inst  & GSM8K  & 400 & 97.5 & 93.2 & 4.3  & [2.8, 5.8]      & $<.001$\rlap{$^{***}$} \\
GPT-4o-mini     & GSM8K  & 400 & 92.5 & 90.2 & 2.2  & [0.5, 4.1]      & .003$^{**}$ \\
DeepSeek-V3     & GSM8K  & 400 & 95.8 & 93.7 & 2.0  & [0.5, 3.5]      & .003$^{**}$ \\
\midrule
Llama-1B-Inst   & ARC    & 200 & 51.5 & 49.4 & 2.1  & [$-$4.2, 8.3]   & .251 \\
Llama-3B-Inst   & ARC    & 200 & 80.5 & 76.4 & 4.1  & [$-$0.5, 8.5]   & .002$^{**}$ \\
Qwen-3B-Inst    & ARC    & 200 & 86.5 & 80.5 & 6.0  & [2.0, 10.0]     & $<.001$\rlap{$^{***}$} \\
Llama-8B-Inst   & ARC    & 200 & 90.5 & 86.2 & 4.2  & [0.3, 7.9]      & $<.001$\rlap{$^{***}$} \\
Qwen-7B-Inst    & ARC    & 200 & 93.5 & 89.1 & 4.4  & [1.8, 7.1]      & $<.001$\rlap{$^{***}$} \\
Mistral-7B-Inst & ARC    & 200 & 68.5 & 63.0 & 5.5  & [1.1, 9.9]      & $<.001$\rlap{$^{***}$} \\
Gemma-2B-it     & ARC    & 200 & 81.5 & 75.2 & 6.2  & [2.0, 10.4]     & $<.001$\rlap{$^{***}$} \\
Gemma-9B-it     & ARC    & 200 & 92.5 & 89.1 & 3.4  & [0.1, 6.7]      & $<.001$\rlap{$^{***}$} \\
Gemma-27B-it    & ARC    & 200 & 93.0 & 87.1 & 5.9  & [2.3, 9.2]      & $<.001$\rlap{$^{***}$} \\
Mistral-Sm-Inst & ARC    & 200 & 94.0 & 91.2 & 2.8  & [0.2, 5.3]      & $<.001$\rlap{$^{***}$} \\
Qwen-72B-Inst   & ARC    & 200 & 96.0 & 93.0 & 3.0  & [1.2, 4.8]      & $<.001$\rlap{$^{***}$} \\
Llama-70B-Inst  & ARC    & 200 & 97.5 & 94.8 & 2.7  & [1.2, 4.2]      & $<.001$\rlap{$^{***}$} \\
GPT-4o-mini     & ARC    & 200 & 97.0 & 94.1 & 2.9  & [1.1, 4.8]      & $<.001$\rlap{$^{***}$} \\
DeepSeek-V3     & ARC    & 200 & 97.5 & 95.2 & 2.3  & [1.1, 3.7]      & $<.001$\rlap{$^{***}$} \\
\midrule
\multicolumn{7}{l}{\textit{MultiArith omitted for space (5/14 open-source models significant; ceiling effects suppress drops).}} \\
\bottomrule
\end{tabular}
\caption{\textbf{Bootstrap confidence intervals for the O$\to$E accuracy drop} (zero-shot CoT prompting) on GSM8K and ARC-Challenge. Thirteen of fourteen models show significant degradation on each of GSM8K and ARC-Challenge (Llama-1B excepted on both due to its low baseline). MultiArith omitted (5/14 significant due to ceiling effects). Additional frontier models (GPT-4o, GPT-5, GPT-5.4, o3) show consistent 1--3\% drops but are not statistically significant at $n{=}200$--400 due to ceiling effects ($>$94\% baseline). Drops are computed from unrounded accuracies and may differ from the printed O$-$E by 0.1.}
\label{tab:bootstrap_ci}
\end{table*}

\end{document}